\pdfoutput=1

\documentclass[11pt]{article}

\usepackage[final]{acl}

\usepackage{times}
\usepackage{latexsym}

\usepackage[T1]{fontenc}

\usepackage[utf8]{inputenc}

\usepackage{microtype}

\usepackage{inconsolata}

\usepackage{graphicx}

\usepackage{subcaption}

\usepackage{booktabs}
\usepackage{dingbat}
\usepackage{stackengine}
\usepackage{tabularx}
\usepackage{tikz}
\usepackage{multirow}
\usepackage{xspace}
\usetikzlibrary{backgrounds}
\usetikzlibrary{fit}
\usetikzlibrary{calc}
\usetikzlibrary{matrix}
\usetikzlibrary{shapes.geometric}
\usepackage{tcolorbox}

\newcommand{\eb}{\mathbf{e}}

\newcommand{\attribute}[1]{\texttt{\spaceskip=0.3em plus 0.1em minus 0.1em #1}}
\newcommand{\concept}[1]{\textsc{#1}}

\newcommand{\things}{\textsc{things}\xspace}
\newcommand{\mxt}{McRae$\times$\textsc{things}\xspace}

\title{Seeing What Tastes Good: \\Revisiting Multimodal Distributional Semantics\\ in the Billion Parameter Era}

\author{Dan Oneata$^*$ \ \ Desmond Elliott$^{\dagger, \ddagger}$ \ \ Stella Frank$^{\ddagger,\dagger}$ \\
$^*$\textsc{politehnica} Bucharest \ \ $^{\ddagger}$Pioneer Center for AI \\
$^{\dagger}$Department of Computer Science, University of Copenhagen\\ 
  \texttt{dan.oneata@gmail.com} \ \ \texttt{stfr@di.ku.dk} \\}

\begin{document}
\maketitle
\begin{abstract}

Human learning and conceptual representation is grounded in sensorimotor experience, in contrast to state-of-the-art foundation models.
In this paper, we investigate how well such large-scale models, trained on vast quantities of data,
represent the \textit{semantic feature norms} of concrete object concepts, e.g.\ a \concept{rose} \attribute{is red}, \attribute{smells sweet}, and \attribute{is a flower}.
More specifically, we use probing tasks to test which properties of objects these models are aware of.
We evaluate image encoders trained on image data alone, as well as multimodally-trained image encoders and language-only models,
on predicting an extended %
denser version
of the classic McRae norms and the newer Binder dataset of attribute ratings.
We find that multimodal image encoders slightly outperform language-only approaches, and that image-only encoders perform comparably to the language models, even on non-visual attributes that are classified as ``encyclopedic'' or ``function''.
These results offer new insights into what can be learned from pure unimodal learning, and the complementarity of the modalities.%
\footnote{Code, datasets and results are available at: \url{https://danoneata.github.io/seeing-what-tastes-good}.}

\end{abstract}

\section{Introduction}

\begin{figure}[t]
  \centering
  \includegraphics[width=0.9\columnwidth]{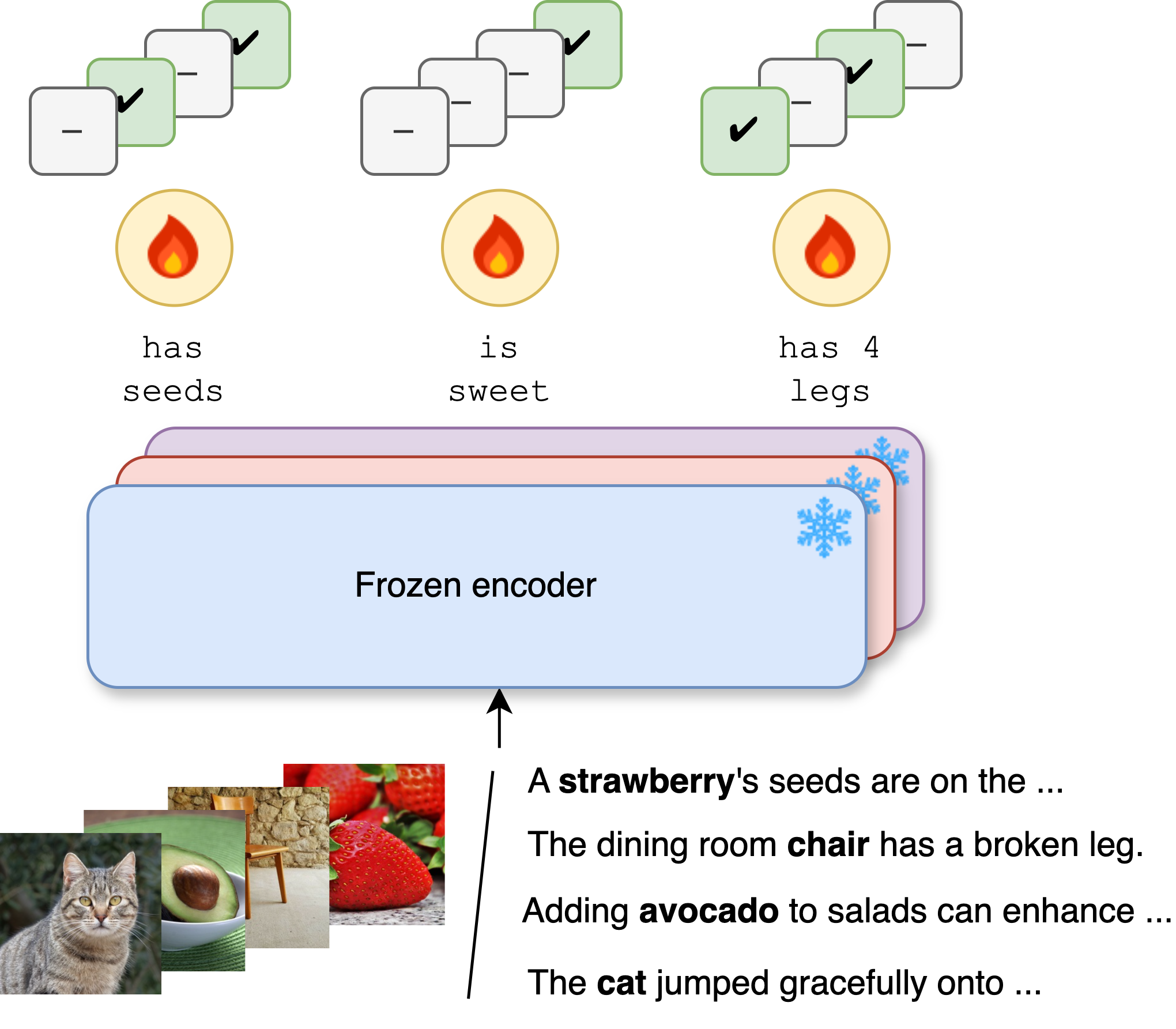}
  \caption{Given a dataset of concrete concepts, depicted using either visual or linguistic data, that are paired with semantic norms, we train linear probes on frozen modality-specific representations of to understand how well conceptual attributes can be extracted from models.}
  \label{fig:methodology_overview}
\end{figure}

Multimodal models depend on vision encoders to provide information about the objects that are depicted, including their properties, spatial configuration, lighting, and scene information.
Recent work has highlighted a degree of linear alignment between neural network representations of the vision and language modalities~\cite{abdou2021,merullo2023,li2024c,huh2024}. 
This implies that the respective representation spaces have similar configurations, in terms of the local organisation (nearest neighbours) of concepts.
However, there remains an open the question of \emph{how} the different modalities  ``understand'' or represent the concepts: which attributes are salient for a  concept? 
In other words, how similar, in terms of underlying attributes: is a \concept{chair} as seen by a vision encoder similar to a \concept{chair} as encoded by a language model?
This question concerns the complementarity of vision and language:
are different modalities distinct, or in fact convergent~\cite{huh2024}?
Is a single modality, such as text, sufficient, or are multiple knowledge sources necessary?
Early work on distributional representations, in text-only~\cite{baroni2008,rubinstein2015,lucy2017,forbes2019neural,misra2022,misra2023} and multimodal~\cite{bruni2014,collell2016} models of static word embeddings, studied this question extensively.
Recent advances in representation learning calls for revisiting this question to understand the relative representational power of each modality in modern models.

In this paper, we investigate how vision, language, and vision-and-language models represent concrete object concepts in terms of their associated attributes (semantic norms).
We use a linear probing methodology to test whether model representations make distinctions corresponding to attributes associated with concepts, depicted visually or in text.
Figure~\ref{fig:methodology_overview} presents an overview of our approach.
The semantic norms cover many types of attributes, from visual-perceptual \attribute{is green}, to the functional \attribute{is eaten}, to the encyclopedic \attribute{grows on trees}. 
Our first question is whether different encoders, from different modalities, capture particular attribute types more or less well.

Secondly, the models we evaluate correspond to a set of hypotheses about the role of language and labelling in conceptualization and category learning, a hotly debated topic in cognitive and neuroscience
\cite{waxman1995,lupyan2012,ivanova2020,benn2023}. %
At one extreme are pure vision encoders (ViT-MAE, DINOv2) trained without any language or category label supervision.
At the other, models like CLIP and SigLIP learn to represent the visual input by aligning it to text as batch-wise nearest neighbours: a form of language-steered world learning.
We also evaluate text-only models that get categories for free (via word labels) but have to infer perceptual and other attributes from distributional semantics.
Inasmuch language ``carves up the world'', visual encoders with more language input should be better aligned with semantic norms for English concepts.

We test these hypotheses using two concept attribute datasets.
The first dataset %
links the semantic norms from the McRae dataset~\cite{mcrae2005} to the concepts of the \textsc{things} project~\cite{hansen2022}, with an additional expansion step, to create the new \mxt dataset.
The second is a dataset of neuro-cognitive attribute ratings from~\citet{binder2016}, which has been used to investigate language model representations~\cite{utsumi2018,turton2020,chronis2023}, but not, to our knowledge, visual or multimodal representations.

Our results demonstrate strong conceptual awareness in multimodal visual encoders across all attribute types.
Moreover, while single-modality models behave most similarly (i.e.~vision models and language models correlate most strongly within-modality), all performant models are highly correlated, indicating a degree of convergence, given exposure to sufficient data of either modality.

The main contributions of this paper include:
\begin{itemize}
\item Improved understanding of the conceptual knowledge embedded in vision encoder models, ranging from self-supervised to class-supervised and language-supervised.
\item \textbf{\mxt}: a new dataset of concepts densely annotated with semantic norms, using attributes from the McRae dataset and concepts from \things.
\item Best practices for extracting representations for lexical semantic probing from LLMs.
\end{itemize}

\section{Related Work}

Understanding and evaluating the lexical semantics learned by language models via co-occurrence patterns is a long-standing concern in distributional semantics. %
A popular method for evaluating vector representations of lexemes is the correlation between the cosine similarity of two words in model space compared to human ratings of word similarity (e.g.~using MEN~\cite{bruni2014} and SimLex~\cite{hill2015}).
However, cosine similarity cannot uncover the underlying dimensions of meaning space, or how the space distinguishes between human-meaningful attributes.
In contrast, testing for specific semantic attributes, by predicting semantic norms, can inform us about the underlying organisation of a model's representation space.

\citet{baroni2008} were the first to use the McRae semantic norm dataset to evaluate the correspondence between early models of distributional semantics and cognitive concepts, using nearest-neighbors procedures.
Using prediction models similar to linear probing, \citet{rubinstein2015,lucy2017} find that static word embeddings encode taxonomic properties significantly more accurately than other types of properties, a finding we replicate.
However, \citet{sommerauer2018} find that embeddings also reliably encode attributes that cut across taxonomic classes, such as \attribute{is dangerous}. 
\citet{fagarasan2015} show that semantic norms can be predicted from word embeddings for unseen concepts.  %
Contextual representation models outperform static word embeddings~\cite{forbes2019neural,bhatia2024b}.
\citet{misra2022,misra2023} use semantic norms to explicitly probe for taxonomic generalization across concepts.

Conceptual attributes (either in the form of McRae norms directly or very similar data) have also been used to investigate the complementarity of representations learned from language and vision.
While~\citet{silberer2013,derby2018a,derby2022} show that multimodal representations can improve norm prediction, i.e. that two modalities are better than one, \citet{bruni2014,collell2016} find only slight patterns of differences when they examine the differences between vision and language representations in predicting different attribute types.  %

This latter finding (which we confirm for more recent models) is in line with more recent work by \citet{merullo2023,li2024c}
which posits a linear relationship between vision and language encodings.
These works also compare across different vision architectures with more or less supervision.
\citet{merullo2023} connect frozen visual encoders to frozen language models with a trained linear transform, and
find that the performance on image captioning correlates with the amount of language supervision of the visual encoder: CLIP, trained with full captions, performs better than a model trained on category labels, and self-supervised BEiT, trained on image data alone, performs worst.
Alternatively, \citet{li2024c} perform Procrustes analysis (a linear mapping) between image representations from ImageNet-trained vision models and language model representations for the same concepts, and find better alignment with larger language models, and with vision models that have been trained on supervised classification tasks, rather than self-supervised learning.

There is less work on the semantic alignment of vision model representations with human conceptual knowledge.
In the computer vision literature, \citet{muttenthaler2023a,mahner2024} has investigated the alignment between vision model representation spaces and human visual similarity judgements, using the \things dataset \cite{hebart2023}.
This work is directly analogous to evaluating pairwise similarities of language model representations against semantic similarity judgements, and as such, doesn't separate out individual concept attributes.
Moreover, it assesses representations corresponding to instances (single images), rather than concepts (collections of instances).
\citet{mahner2024} compare sparse representations of human and model similarities, finding that while core dimensions overlap, humans use more semantic cues, and vision models rely more on visual cues, as well as many human-uninterpretable cues.  %
In a study of several vision encoders, \citet{muttenthaler2023a} find that models trained on larger datasets and language supervision (CLIP) are more aligned with human similarity than smaller label- and self-supervised models.
Finally, \citet{suresh2024} show that image encoders trained to predict object attributes, rather than object classes, are more aligned with humans.

\section{Concept Attributes: Datasets}  %
\label{subsec:concept-norms}

Understanding concepts via a core set of distinctive attributes is a long-standing quest in cognitive science~\cite{aristotle1928,rosch1975,nosofsky2018a,gardenfors2000}.
One method of discovering which attributes are important for human categorisation is \emph{semantic norm elicitation}: participants are asked to write down the ``characteristics and attributes'' \cite{rosch1975} or ``properties'' \cite{mcrae2005} they associate with a particular concept.
Pooled over many participants, semantic norms thus represent a concept as a set of frequently mentioned salient attributes.

While commonly used, semantic norm data have two important weaknesses.
Firstly, they are not \emph{complete}: less-salient, but nonetheless present, attributes of concepts are often missing (e.g.~\concept{tiger} but not \concept{cat} \attribute{has teeth}).
To remedy this first issue, we synthetically ``complete'' the attribute values from \citep{mcrae2005} across a large set of concepts.
Secondly, norms are \emph{biased} towards attributes that are easily lexicalised.
We thus also explore
a recent dataset of ratings across a fixed set of attributes related to sensory and neurological dimensions
that are not based on elicited lexicalised norms ~\cite{binder2016}.

Since we are exploring visual and linguistic representations, the concepts we consider are concrete objects, corresponding to English nouns.
We use the set of object concepts from \things~\cite{hebart2019}, which also includes a set of quality-controlled images for each concept.

\paragraph{\mxt norms.}
The original McRae semantic norms dataset~\cite{mcrae2005} contains 541 concepts and 2\,524 unique norms.
The attributes are classified into different types, such as `taxonomic', `functional', `visual-colour', corresponding to associated brain regions~\cite{cree2003}.
We discard attributes appearing with fewer than five concepts;
we also group highly similar attributes
(e.g.~\attribute{used by the military}, \attribute{used by soldiers}, \attribute{used by the army}) using 
semantic similarity.%
\footnote{%
We merge attributes with cosine similarity greater than 0.9, using the sentence embedding model
\href{https://huggingface.co/sentence-transformers/all-MiniLM-L6-v2}{all-MiniLM-L6-v2}.}
This results in a final set of 278 attributes.
We then find the corresponding norms/attribute values for all 1\,854 concepts in \things, resulting in a densely annotated dataset without missing norms.

To obtain a complete mapping between concepts and attributes, 
we ask GPT-4o to annotate whether or not each attribute is a common trait of each concept (see Appendix~\ref{sec:appendix-gpt4o-dataset});
each concept is briefly disambiguated and described using a definition extracted from the \things metadata.
As a sanity check we verify that the norms (concept--attribute pairs) produced by our method include the norms in the original McRae set.
We obtain a recall of 94--100\% at responding correctly with respect to the human-authored attributes for a selection of ten attributes (one for each category),
and, as desired, the number of concepts positively associated with a given attribute increases. 
For example, the number of positive concepts for \attribute{tastes good} increases from 28 to 335; 
for \attribute{lays eggs} from 39 to 83;
for \attribute{is dangerous} from 121 to 299.

We note that \citet{hansen2022} also used an LLM-based process to collect norms for \things,
but their process was designed to elicit more (potentially unique) norms for these concepts, whereas ours has the goal of comprehensive attribute annotation to avoid false negatives (missing positive values).

\paragraph{Binder ratings.}
\citet{binder2016} collected dense ratings for 65 ``experiential attributes'' of 534 concepts, of which we use the 155 concepts also found in \things.
The experiential attributes correspond to lower-level conceptual dimensions such as visual \attribute{brightness}, somatic \attribute{pain}, or motor \attribute{movements in the upper/lower body}, and are organized into 14 different fine-grained domains (vision, somatic, etc.), collapsed to 7 coarser domains (sensory, motor,~etc.).
Participants used a 7-level rating scale%
\footnote{They answered the question ``To what degree do you think of \textsc{concept} as having/being associated with \textsc{attribute}?''}
and the final concept-attribute rating is the mean across participants.

\begin{table*}[t]
    \centering
    \setlength{\tabcolsep}{4pt}
    \begin{tabular}{lllllll}
         \toprule
         Model & Params. & Dataset & Size & Objective & Labels & IN-1K\\
         \midrule
         FastText & -- & CommonCrawl & 840B & NLL & -- & --\\
         GLoVe & -- & CommonCrawl & 840B & NLL & -- & -- \\
         Numberbatch & -- & ConceptNet & N/A$^\ddagger$ & PPMI & -- & -- \\
         DeBERTa~v3 & 86M & Wiki+Books & 3.1B & RTD & -- & -- \\
         Gemma & 2B & Private & 6T & NLL & -- & --\\
         \midrule
         ViT-MAE & 304M & ImageNet-1K & 1.3M & MSE & N/A & 85.9\\
         Max ViT (IN-1K)$^\dagger$  & 212M & ImageNet-1K & 1.3M & Classification & Object classes & 85.2 \\
         Max ViT (IN-21K) & 212M & ImageNet-21K & 14M & Classification & Object classes & 88.3 \\
         Swin-V2$^\dagger$ & 197M & ImageNet-21K & 14M &SimMIM & N/A & 87.7\\
         DINOv2 & 304M & LVD & 142M & DINO + iBOT & N/A & 86.3\\
         \midrule
         CLIP & 304M & Private & 400M & Contrastive & Sentences & 83.9 \\
         SigLIP & 400M & Private & 4B & Sigmoid Contr. & Sentences & 83.2 \\
         PaliGemma & 400M & Private & 1B & NLL & Sentences & N/A \\
         LLaVa-1.5 &  324M & CC3M, OKVQA, etc. & 1.2M & NLL & Sentences & N/A \\
         Qwen2.5-VL & 669M & Private & UNK & NLL & Sentences & N/A \\
         \bottomrule
         & 
    \end{tabular}
    \caption{Overview of the models studied in this paper. The number of parameters in the encoder, the type and size of the pretraining data, the pretraining objective, and, where applicable, the reported \textbf{I}mage\textbf{N}et\textbf{1K} classification accuracy at 224px $\times$ 224px, except where noted otherwise.
    $\dagger$: 384px $\times$ 384px. $\ddagger$: ConceptNet is a knowledge graph of words and phrases with 8M nodes and 21M edges.
    }
    \label{tab:models-overview}
\end{table*}

\section{Models}
\label{subsec:models}

We primarily study the performance of
image encoder models using  Vision Transformers (ViT) backbones~\cite{dosovitskiy2020image}, trained with different amounts of linguistic supervision.
Table~\ref{tab:models-overview} presents a high-level overview.
At one extreme, we use visual encoders trained \emph{without any} label supervision. We also use encoders trained with object label classification supervision, e.g.\ trained on the ImageNet dataset. 
At the other end of the spectrum, we use visual encoders resulting from large-scale vision-language contrastive learning, and encoders derived from vision-language generative pretraining. 
The models were chosen so the encoders are approximately the same size, and operate over the same patch sizes. 
We also evaluate text-only embedding models,
to compare the conceptual knowledge learned from the textual modality.
Appendix~\ref{app:model-details} Table~\ref{tab:model_names} shows the exact model names used in \texttt{timm} / HuggingFace Transformers.

\subsection{Vision-only Models}
\label{subsubsec:vision-only}

\textbf{ViT-MAE}~\cite{he2022masked} is a self-supervised visual encoder pre-trained to reconstruct masked image patches at the pixel level using a deep Transformer encoder and decoder. 
\textbf{DINOv2}~\cite{oquabdinov2} is also a self-supervised visual encoder pretrained using a combination of image-level objectives and patch-level objectives using a student and a teacher network~\cite{moutakanni2024}. 
This model is trained on a very large diverse dataset (142M images) without labels.
\textbf{Swin-V2}~\cite{liu2022swin} is a self-supervised visual encoder pretrained on ImageNet-21K to reconstruct masked image patches using a single linear layer~\cite{xie2022simmim}.
\textbf{Max ViT}~\cite{tu2022maxvit} is a Vision Transformer with Transformer blocks that combine convolution, block attention, and grid-based attention.
This model is directly trained with a multi-class classification objective on 
ImageNet (IN-1K or IN-21K).

\subsection{Multimodal Models}

\textbf{CLIP}~\cite{radford2021learning} has separate visual and textual encoders that are jointly optimized to maximize the similarity of image--sentence pairs. 
\textbf{SigLIP}~\cite{zhai2023sigmoid} also has separate encoders that are trained to maximize a compute-efficient contrastive sigmoid loss. 
\textbf{PaliGemma}~\cite{beyer2024paligemma} is a generative vision-language model initialized from the SigLIP-So400M visual encoder and the Gemma language model~\cite{team2024gemma}. 
It is then further trained on a multimodal conditional language modelling task, and
we use the visual encoder at the end of this multi-stage multimodal pretraining. 
\textbf{LLaVa-1.5}~\cite{liu2024llava} is also a generative model that projects CLIP ViT/L embeddings into the Vicuna-7B language model \cite{zheng2023vicuna} using an MLP projector.
The model is multimodally trained on instruction data generated with GPT-4 on the CC3M dataset \cite{sharma2018conceptual},
as well as on other scientific visual question answering datasets.
\textbf{Qwen2.5-VL}~\cite{bai2025qwen2} similarly integrates vision information through projection into an large language model,
but in this model the image is input as a series of tokens, rather than as a single embedding.
The model is trained in multiple stages on a wide variety of proprietary multimodal data.

\subsection{Language-only Models}

\textbf{FastText}~\cite{mikolov2018advances} creates static word embeddings by combining character n-grams embeddings within a white space-delimited word. 
\textbf{GLoVe}~\cite{pennington2014glove} also creates static embeddings based on aggregated global word-word co-occurrence statistics. 
For both FastText and GLoVe we use 300D embeddings trained on Common Crawl (840B tokens).
\textbf{Numberbatch}~\cite{speer2017conceptnet} embeddings (300D) are a combination of ConceptNet graph embeddings plus GLoVe and word2vec embeddings.
\textbf{Gemma}~\cite{team2024gemma} is a 2B parameter causal language model trained on 3T tokens.
\textbf{DeBERTa~v3} is an language encoder trained on Wikipedia and the Books Corpus (3.1B words) to detect replaced tokens in sentences. 
\textbf{CLIP}~\cite{radford2021learning} also has a language encoder; we use the 151M parameter model that was trained with the visual encoder.

\section{Methodology}

We use trained linear probes~\cite{alain2017,hupkes2018,belinkov2022probing} to measure
the extent to which conceptual attributes (McRae feature norms or Binder attribute ratings) are evident in image and text representations.
This evaluation requires generalizing attributes to unseen concepts, based on a small set of positive examples.
Following standard methodology, the linear probes are trained on top of frozen representations, which allows us to estimate what is captured in the representations directly.

Each attribute is learned with a separate probe.
For \mxt, we train a linear classifier for each attribute that maps a concept representation to a binary label, using a simple logistic regression.\footnote{We use \texttt{sklearn}'s default implementation without regularization and increase the maximum number of iterations to 1\,000. We cannot train more elaborate (MLP) probes since our training datasets are very small, with few positive examples.}
For the Binder ratings dataset, we train a linear regression on each attribute to predict the mean rating for each concept-attribute pair.%
\footnote{We use the \texttt{LinearRegression} implementation from \texttt{sklearn} with default settings: fit intercept, no regularisation.}
For both datasets, we generate 10 train--test splits for each attribute using 5-fold stratified cross-validation repeated twice, and report the average performance.

\paragraph{Visual concept representations.}
In the visual modality, a concept is represented by images from its \things concept class.
The visual concept $\eb_c$ is computed by averaging the embeddings extracted from the last layer of a given vision encoder.
Since many of the vision models produce a dense grid of embeddings, we obtain a single vector by average pooling the embeddings spatially.

\paragraph{Textual concept embeddings.}
In the language modality, a concept is represented by the English noun label given by McRae.
Static word embedding models (GloVe, FastText, Numberbatch return an embedding directly, using only the surface form of the word.%
\footnote{The static embeddings for multi-word concepts are averaged; homophones are not distinguished.}
Contextual language models (Gemma, DeBERTa~v3) require a more careful methodology to extract meaningful vector representations.
In these results, we always average over 10 sentences of the word in context (collected from the GPT4o API, see Appendix~\ref{sec:appendix-gpt4o-dataset}), following~\citep{vulic2020,bommasani2020}.
We find that each model requires a different extraction technique in order to achieve reasonable performance; see~Appendix~\ref{sec:appendix-textual_contexts} for failed attempts and suggestions for best practices.
Briefly, the best representations are found from mean-pooling over multiple layers~\citep{vulic2020}.
For Gemma, we obtain much better performance using only the last token of the target word, while for the masked language model (DeBERTa~v3) we use the mean over all concept tokens.

\subsection{Evaluation and Baselines}

For \mxt, our main evaluation metric is F$_1$ score.
Following~\cite{hewitt2019}, we calculate the \emph{selectivity} of each probe as the difference between the F$_1$ score on the correct labelling minus the expected random performance (i.e.~the expected performance of a probe that learned a frequency bias).
F$_1$ selectivity results are thus already with regard to a random baseline.
(A second random baseline is provided by the \textbf{SigLIP-Random} encoder, an untrained, randomly initialized, version of SigLIP.)
For the linear regression results on Binder, we report root mean squared error (RMSE) as the main metric.
(We also include F$_1$ accuracy results for logistic regression on a median-binarised version of Binder in Appendix~\ref{app:further-results}.)

\section{Results}

\begin{table}[t]
    \centering
    
\newcommand\annot[1]{\color{gray}\scriptsize{#1}}
\setlength{\tabcolsep}{2pt}

\begin{tabularx}{\linewidth}{Xcc}

\toprule
      & \multicolumn{1}{c}{\stackon{\things}{McRae$\times$}} & \multicolumn{1}{c}{Binder} \\
Model & \multicolumn{1}{c}{\color{gray} \small F$_1$ sel $\uparrow$} & \multicolumn{1}{c}{\color{gray} \small RMSE $\downarrow$} \\
\midrule
\multicolumn{3}{l}{\textit{Vision models}} \\
Random SigLIP    & 15.4       & 1.43     \\
ViT-MAE          & 35.6       & 0.94     \\
Max ViT (IN-1K)  & 29.0       & 1.37     \\
Max ViT (IN-21K) & 43.3       & 0.84     \\
DINOv2           & 44.5       & 0.80     \\
Swin-V2          & \bf 47.0   & \bf 0.74 \\
\midrule
\multicolumn{3}{l}{\textit{Multimodal vision models}} \\
LLaVA-1.5        & 45.0      & 0.83       \\
Qwen2.5-VL       & 46.8      & 0.79       \\
CLIP (image)     & 48.4      & 0.74       \\
PaliGemma        & 49.9      & 0.73       \\
SigLIP           & \bf 50.1  & \bf 0.71   \\
\midrule
\multicolumn{3}{l}{\textit{Language models}} \\
GloVe 840B       & 39.1       & 0.89        \\
FastText         & 40.2       & 0.91        \\
Numberbatch      & 44.1       & 0.83        \\
CLIP (text)      & 43.0       & 0.81        \\
DeBERTa v3       & 45.5       & 0.68        \\
Gemma            & \bf 49.8   & \bf 0.67    \\
\bottomrule
\end{tabularx}
    \caption{%
        Performance of linear probes, averaged across attributes, for semantic norms on \mxt, and concept attribute ratings on Binder.
        We report F$_1$ selectivity on \mxt, which is corrected for random performance.
        On Binder we perform linear regression and report the root mean squared error (RMSE).
        More results can be found in Appendix Table~\ref{tab:main-results-full}.
    }
    \label{tab:main-results}
\end{table}

\subsection{Main Results}
\label{subsec:main-results}

\begin{figure*}[t]
    \centering
    \includegraphics[width=0.95\linewidth]{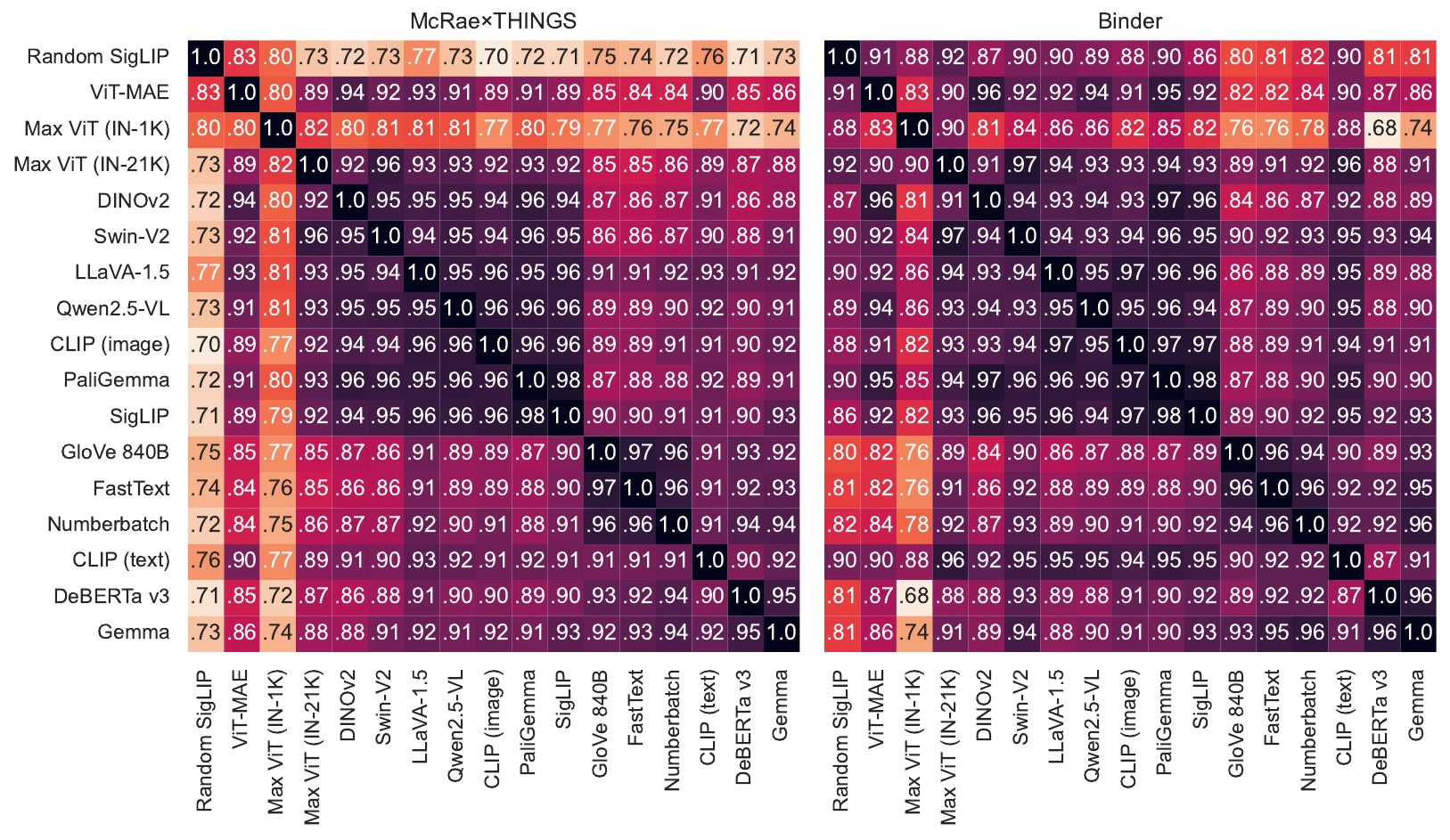}
    \caption{%
    Per-attribute
        Pearson correlation between models on \mxt and Binder datasets.
    }
    \label{fig:correlation between models}
\end{figure*}

\begin{figure}
    \centering
    \includegraphics[width=0.87\linewidth]{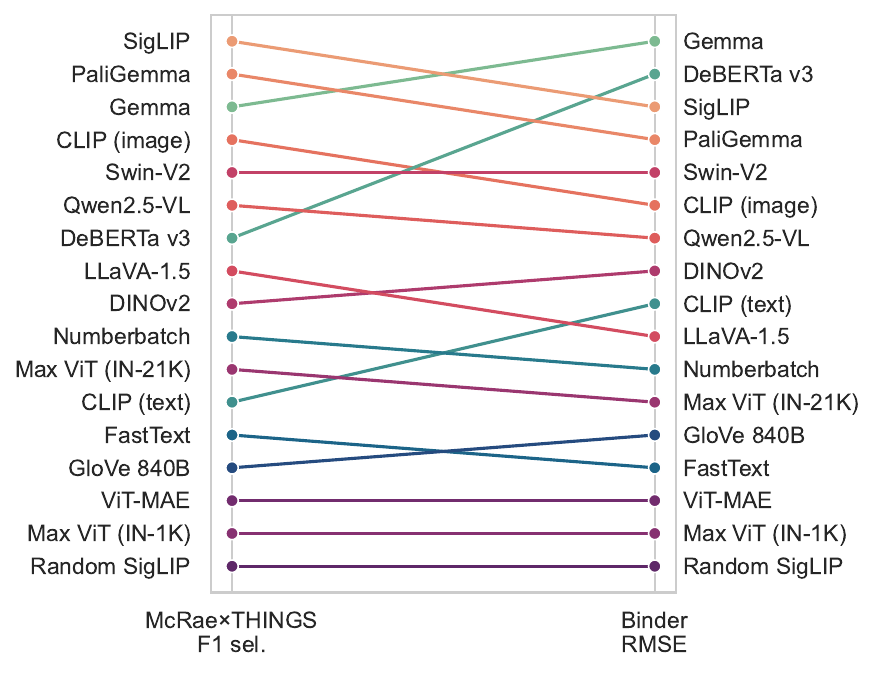}
    \caption{
        Relative rankings of models across the \mxt and Binder datasets (higher rank is better).
        The vision models are show in warm colours, language models in cool colours. }
\label{fig:rankings}
\end{figure}

The results for linear probe accuracy results are shown in Table~\ref{tab:main-results}; see also Appendix~\ref{app:further-results}, Table~\ref{tab:main-results-full}.

\paragraph{The impact of modality.}
Across the two datasets, the multimodal vision encoders
are consistently amongst the highest performing models.
However, the large text-only LLMs (Gemma-2B and DeBERTa~v3)
also rank highly.
The self-supervised Swin-V2 model is the best vision model, 
and clearly outperforms (among others) the static word embedding models, despite having no access to lexical information.

\paragraph{Dataset differences.}
Text-only models (especially Gemma and DeBERTa~v3) perform relatively better on the Binder attribute dimensions, as seen in the rankings (Figure~\ref{fig:rankings}), while visually-informed models predict \mxt attributes better.
Both dataset show large variation across different attributes.

\paragraph{Effect of training data amounts.}  %
Language models trained on larger amounts of data perform consistently better on \mxt and Binder.
On the vision side,
Swin-V2 learns better representations than DINOv2 for predicting semantic attributes, despite having seen one tenth as much data (14M vs 142M).
Swin-V2 also outperforms the label-supervised ViT-MAE (IN-21K), having been trained on the same dataset, but with a less-informed objective.
However, for Max ViT, the training data size has a substantial impact.
For the multimodal vision models, the results on \mxt suggest that training data matters to some degree; for example, CLIP (image)%
\footnote{%
We also evaluated the vision encoder of the performant open-weight (\href{https://huggingface.co/apple/DFN2B-CLIP-ViT-L-14}{DFN2B-CLIP-ViT-L-14}) trained on DataComp-1B for the same number of total training examples as OpenAI CLIP.
It achieves an F$_1$ selectivity of 47.7 for \mxt and an RMSE of 0.79 on Binder.}
(400M) is outperformed by SigLIP and PaliGemma (5B).
However, it is hard to disentangle the effect of dataset size from architecture and, in the case of language models, probing methodology (see Appendix~\ref{sec:appendix-textual_contexts}).

\begin{figure*}[t]
    \centering
    \includegraphics[width=1\linewidth]{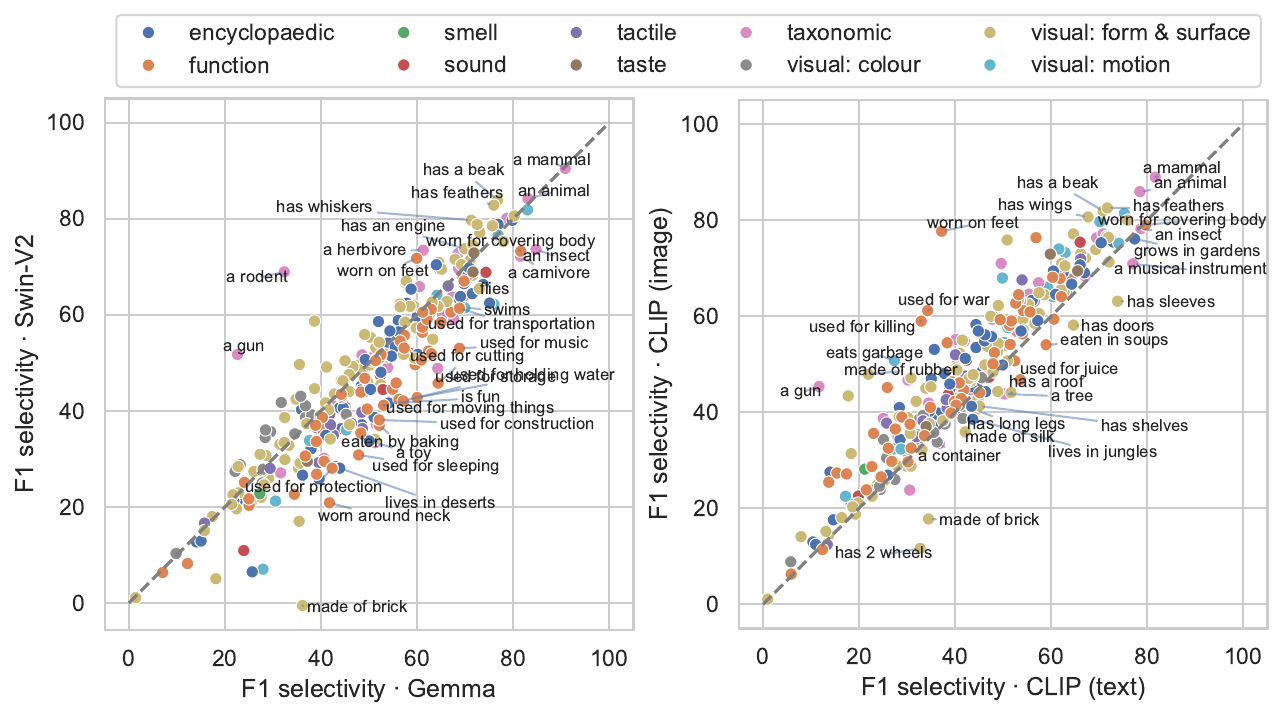}
    \caption{%
        Per feature comparison between pairs of models in terms of the F1 selectivity score.
        Left: Swin-V2 vs Gemma.
        Right: CLIP (image) vs CLIP (text).
    }
    \label{fig:fasttext-vs-siglip}
\end{figure*}

\paragraph{Correlation between model predictions.}
To understand the difference in model behaviour at the level of individual attributes, we calculate pairwise Pearson correlations between probe accuracy on different models (Figure~\ref{fig:correlation between models}).
For the \mxt norms and Binder attributes, we see modality clusters, where vision encoders (with the exception of Max ViT IN-1K) are correlated with each other, and likewise the static word embedding models and the LLMS Gemma and DeBERTa~v3.
We also see some cross-modal correlations, with CLIP (text) correlating relatively higher with vision models in general (not only the CLIP image encoder), and Swin-V2 correlating more highly with language models on Binder.
Overall, all (reasonable) model correlations are quite high, indicating that good encoders across modalities are rather similar.
Inspecting the best and worst attribute for each model shows high consistency: For \mxt, the most accurate attribute across models is \attribute{is mammal}, while the worst is consistently \attribute{different sizes}.\footnote{Interestingly, this is an attribute that is clearly associated with the (variation shown by the) concept, instead of being associated with individual instances.}
For Binder the easiest attribute is \attribute{angry}, while the hardest is \attribute{sound} for most models.
Figure~\ref{fig:fasttext-vs-siglip} visualizes norm prediction performance of specific pairs of models (vision-only Swin-V2 vs text-only Gemma, CLIP image vs CLIP text), and qualitative examples can be found in Appendix~\ref{sec:qualitiative results}.

\subsection{Attribute Type Results} %

\begin{figure*}
    \centering
    \includegraphics[width=0.99\linewidth]{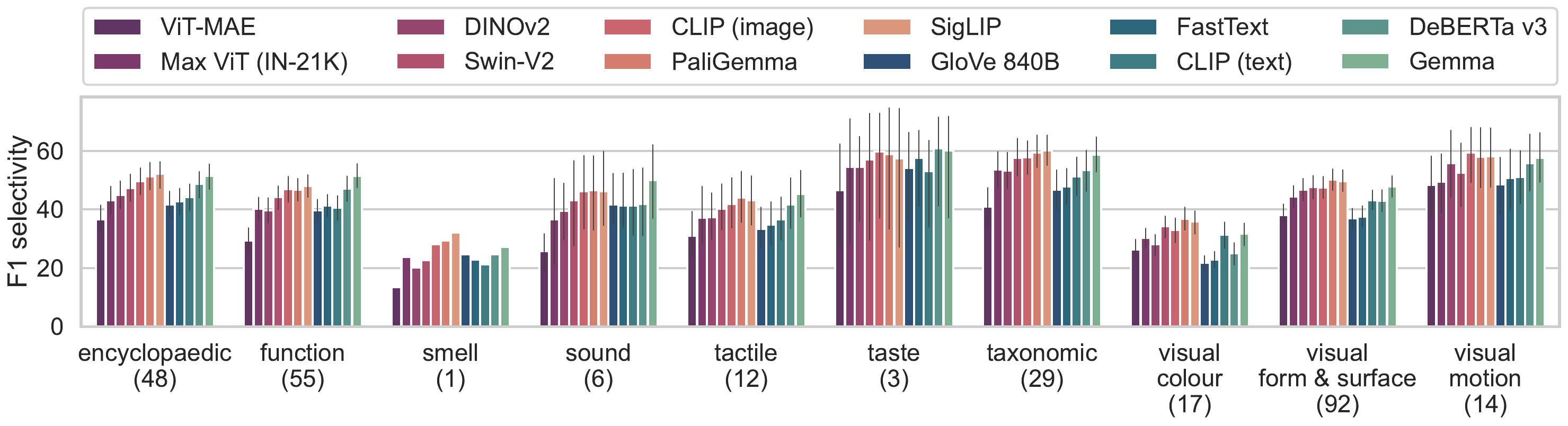}
    \caption{%
        Results (F$_1$ selectivity) per attribute (norm) type on the \mxt data. 
        The number below each type indicates the number of norms belonging to that type.
        The error bars denote 95\% confidence intervals using bootstrapping.
        Vision models are in reddish colours, while language models are in greenish colours.
    }
    \label{fig:per-metacategory-mcrae-extended}
\end{figure*}

Are vision encoders better at visual-perceptual features?
Do language models encode more functional-encyclopedic features?
To answer these questions we study performance aggregated by attribute type, as given by the datasets.
Figure~\ref{fig:per-metacategory-mcrae-extended} presents the \mxt probing results per attribute type.
Among the ten types,
we see that taxonomic, visual-motion, and taste attributes are the easiest to predict.
The vision models, especially the multimodal models, generally outperform the static word embeddings and to some extent the language models (Gemma and DeBERTa~v3).
This makes sense for visual attributes like colour, but,
surprisingly, this is the case even for ``encyclopedic'' and ``functional'' attributes, which should be easier to learn from text than from visual inputs.
Results by Binder attribute domain 
(Appendix~Figs.~\ref{fig:binder-norms-results} and~\ref{fig:binder-norms-results-by-type})
show similar patterns, with strong LLMs, multimodal vision encoders, and Swin-V2 performing similarly across attribute domains.

\paragraph{Possible confounds.}
Since linear probes are learned using attribute extensions (the set of positive examples of an attribute), we cannot be sure they actually learn the attribute characteristics, and not some closely correlated, but more visually or textually available, attribute.
For example, the two taste attributes (\attribute{tastes good} and \attribute{tastes sweet}) have extensions that are subsets of the food supercategory, which is learnable from visual features alone (e.g.~as demonstrated by high performance on the taxonomic \attribute{is food} norm for all models).
Likewise, many of the motion attributes capture subsets of animals (\attribute{eats grass}).
As a initial analysis, we check whether models are better at learning attributes that coincide with taxonomic supercategories, as provided by the \things dataset.
The resulting correlations (Table~\ref{tab:tax-norm-corr}) are highest for 
CLIP-image (0.594), FastText (0.578), and Numberbatch (0.573),
a heterogenous set of models in terms of modality and their linear probing accuracy.

\begin{table}
    \centering
    \begin{tabularx}{\linewidth}{Xlr}
\toprule
Model            & Modality & Correlation \\
\midrule
CLIP (image)     & V($+$L) & 0.594 \\
FastText         & L       & 0.578 \\
Numberbatch      & L       & 0.573 \\
LLaVA-1.5        & V($+$L) & 0.565 \\
GloVe 840B       & L       & 0.564 \\
SigLIP           & V($+$L) & 0.561 \\
PaliGemma        & V($+$L) & 0.554 \\
Qwen2.5-VL       & V($+$L) & 0.553 \\
DINOv2           & V       & 0.552 \\
CLIP (text)      & L($+$V) & 0.550 \\
Swin-V2          & V       & 0.545 \\
Gemma            & L       & 0.543 \\
Max ViT (IN-21K) & V       & 0.542 \\
DeBERTa v3       & L       & 0.536 \\
ViT-MAE          & V       & 0.495 \\
Max ViT (IN-1K)  & V       & 0.413 \\
Random SigLIP    & V       & 0.339 \\
\bottomrule
\end{tabularx}
    \caption{%
        \mxt dataset: Pearson correlation between per-norm probing performance, as measured by F$_1$ selectivity, and the proportion of the norm's extension belonging to a single supercategory (i.e.~the extent to which predicting the supercategory would lead to high precision). 
        Modality indicates which input each model operates on: vision (V) or language (L), with multimodality indicated in brackets.
        }
    \label{tab:tax-norm-corr}
\end{table}

\section{Conclusion}

This linear probing analysis on two datasets shows that multimodally-trained vision encoders represent conceptual attributes better than single-modality vision-only or text-only encoders.
However, the single-modality encoders still perform well.
In particular, the self-supervised Swin-V2, and to a lesser extent DINOv2 models, have learned a large amount of conceptual attribute knowledge, comparable to modern LLMs, and more than static word embeddings. 
This result is particularly surprising given that these vision models have not been trained to distinguish between concepts, rather than instances, at all.
Intriguingly, label-supervision of vision models seems to be harmful for learning human-aligned attributes, judging by the relatively worse performance of Max ViT, trained on ImageNet classification, compared to the self-supervised Swin-V2. 

There is a long-held belief that we need multimodally-grounded representations  to overcome the limitations of learning from only linguistic data.
Our results suggest that Vision and Language encoders encode (somewhat) complementary views of concepts %
inasmuch same-modality models correlate stronger than different-modality models.
However, overall correlations are high, indicating a level of convergence. %
Previous claims of modality convergence have used nearest-neighbours
measures~\citep{huh2024,li2024c}; 
here we show similar convergence results using a very different linear probing methodology.

We expect models with conceptual knowledge organised in human-like ways, that are aware of the semantic attributes that underlie category memberships, would, in turn, achieve better downstream performance in language processing tasks.
In future work, we will investigate the predictive power and utility of our probing tasks for multimodal training.
This will also require going beyond simple object concepts to investigate more abstract, situational and configurational, concepts, in order to cover a larger proportion of the human conceptual repertoire.

\clearpage

\section*{Limitations}

\paragraph{Linear probes} 
Our linear probes assume that semantic attributes are encoded linearly in representation space.
However, it is possible that semantic attributes are encoded as non-linear combinations: \cite{sommerauer2018} see increased probing accuracy with small MLPs compared to a logistic regression model such as we used.
Our datasets are too small to learn MLPs without severe overfitting.

\paragraph{English-only}
Our experiments and analyses only concern evaluating the ability of models to predict the English semantic attributes of concepts expressed in English.
This hinders our ability to make broader claims about the ability of models to perform this task in other languages, or for non-Western concrete concepts~\cite{liu2021visually}.
In future work, we are interested in understanding the degree and quality of English-language influence on visual encoder representations.

\paragraph{Risks} We forsee no risks associated with this research.

\section*{Acknowledgments}
Dan Oneata is supported by the EU Horizon project AI4TRUST (No. 101070190) and by CNCS-UEFISCDI (PN-IV-P8-8.1-PRE-HE-ORG-2023-0078).
Desmond Elliott is supported by a research grant (VIL53122) from VILLUM FONDEN.
Stella Frank is supported by the Pioneer Center for AI, DNRF grant number P1.

\bibliography{custom,stella-visnorms-old}

\begin{thebibliography}{64}
\providecommand{\natexlab}[1]{#1}

\bibitem[{Abdou et~al.(2021)Abdou, Kulmizev, Hershcovich, Frank, Pavlick, and
  S{\o}gaard}]{abdou2021}
Mostafa Abdou, Artur Kulmizev, Daniel Hershcovich, Stella Frank, Ellie Pavlick,
  and Anders S{\o}gaard. 2021.
\newblock Can language models encode perceptual structure without grounding?
  {A} case study in color.
\newblock In \emph{Proc. CoNLL}.

\bibitem[{Alain and Bengio(2017)}]{alain2017}
Guillaume Alain and Yoshua Bengio. 2017.
\newblock Understanding intermediate layers using linear classifier probes.
\newblock In \emph{Proc. ICLR Workshop Track}.

\bibitem[{Aristotle(4th c. BC / 1928)}]{aristotle1928}
Aristotle. 4th c. BC / 1928.
\newblock \emph{Categories (Translated by {{E}}. {{M}}. {{Edghill}})}.

\bibitem[{Bai et~al.(2025)Bai, Chen, Liu, Wang, Ge, Song, Dang, Wang, Wang,
  Tang et~al.}]{bai2025qwen2}
Shuai Bai, Keqin Chen, Xuejing Liu, Jialin Wang, Wenbin Ge, Sibo Song, Kai
  Dang, Peng Wang, Shijie Wang, Jun Tang, et~al. 2025.
\newblock Qwen2.5-{VL} technical report.
\newblock \emph{arXiv preprint arXiv:2502.13923}.

\bibitem[{Baroni and Lenci(2008)}]{baroni2008}
Marco Baroni and Alessandro Lenci. 2008.
\newblock Concepts and properties in word spaces.
\newblock \emph{Italian Journal of Linguistics}, 20(1):55--88.

\bibitem[{Belinkov(2022)}]{belinkov2022probing}
Yonatan Belinkov. 2022.
\newblock Probing classifiers: Promises, shortcomings, and advances.
\newblock \emph{Computational Linguistics}, 48(1):207--219.

\bibitem[{Benn et~al.(2023)Benn, Ivanova, Clark, Mineroff, Seikus, Silva,
  Varley, and Fedorenko}]{benn2023}
Yael Benn, Anna~A Ivanova, Oliver Clark, Zachary Mineroff, Chloe Seikus,
  Jack~Santos Silva, Rosemary Varley, and Evelina Fedorenko. 2023.
\newblock \href {https://doi.org/10.1093/cercor/bhad289} {The language network
  is not engaged in object categorization}.
\newblock \emph{Cerebral Cortex}, 33(19):10380--10400.

\bibitem[{Beyer et~al.(2024)Beyer, Steiner, Pinto, Kolesnikov, Wang, Salz,
  Neumann, Alabdulmohsin, Tschannen, Bugliarello et~al.}]{beyer2024paligemma}
Lucas Beyer, Andreas Steiner, Andr{\'e}~Susano Pinto, Alexander Kolesnikov,
  Xiao Wang, Daniel Salz, Maxim Neumann, Ibrahim Alabdulmohsin, Michael
  Tschannen, Emanuele Bugliarello, et~al. 2024.
\newblock Pali{G}emma: A versatile 3{B} {VLM} for transfer.
\newblock \emph{arXiv preprint arXiv:2407.07726}.

\bibitem[{Bhatia and Richie(2024)}]{bhatia2024b}
Sudeep Bhatia and Russell Richie. 2024.
\newblock \href {https://doi.org/10.1037/rev0000319} {Transformer networks of
  human conceptual knowledge.}
\newblock \emph{Psychological Review}, 131(1):271--306.

\bibitem[{Binder et~al.(2016)Binder, Conant, Humphries, Fernandino, Simons,
  Aguilar, and Desai}]{binder2016}
Jeffrey~R. Binder, Lisa~L. Conant, Colin~J. Humphries, Leonardo Fernandino,
  Stephen~B. Simons, Mario Aguilar, and Rutvik~H. Desai. 2016.
\newblock \href {https://doi.org/10.1080/02643294.2016.1147426} {Toward a
  brain-based componential semantic representation}.
\newblock \emph{Cognitive Neuropsychology}, 33(3-4):130--174.

\bibitem[{Bommasani et~al.(2020)Bommasani, Davis, and Cardie}]{bommasani2020}
Rishi Bommasani, Kelly Davis, and Claire Cardie. 2020.
\newblock \href {https://doi.org/10.18653/v1/2020.acl-main.431} {Interpreting
  pretrained contextualized representations via reductions to static
  embeddings}.
\newblock In \emph{Proc. ACL}.

\bibitem[{Bruni et~al.(2014)Bruni, Tran, and Baroni}]{bruni2014}
E.~Bruni, N.~K. Tran, and M.~Baroni. 2014.
\newblock \href {https://doi.org/10.1613/jair.4135} {Multimodal
  {{Distributional Semantics}}}.
\newblock \emph{Journal of Artificial Intelligence Research}, 49:1--47.

\bibitem[{Chronis et~al.(2023)Chronis, Mahowald, and Erk}]{chronis2023}
Gabriella Chronis, Kyle Mahowald, and Katrin Erk. 2023.
\newblock A method for studying semantic construal in grammatical constructions
  with interpretable contextual embedding spaces.
\newblock In \emph{Proc. ACL}.

\bibitem[{Collell and Moens(2016)}]{collell2016}
Guillem Collell and Marie-Francine Moens. 2016.
\newblock Is an image worth more than a thousand words? {O}n the fine-grain
  semantic differences between visual and linguistic representations.
\newblock In \emph{Proc. COLING}.

\bibitem[{Cree and McRae(2003)}]{cree2003}
George~S. Cree and Ken McRae. 2003.
\newblock \href {https://doi.org/10.1037/0096-3445.132.2.163} {Analyzing the
  factors underlying the structure and computation of the meaning of chipmunk,
  cherry, chisel, cheese, and cello (and many other such concrete nouns).}
\newblock \emph{Journal of Experimental Psychology: General}, 132(2):163--201.

\bibitem[{Derby(2022)}]{derby2022}
Steven Derby. 2022.
\newblock \emph{Interpretable Semantic Representations from Neural Language
  Models and Computer Vision}.
\newblock Ph.D. thesis, Queen's University, Belfast.

\bibitem[{Derby et~al.(2018)Derby, Miller, Murphy, and Devereux}]{derby2018a}
Steven Derby, Paul Miller, Brian Murphy, and Barry Devereux. 2018.
\newblock \href {https://doi.org/10.18653/v1/K18-1026} {Using sparse semantic
  embeddings learned from multimodal text and image data to model human
  conceptual knowledge}.
\newblock In \emph{Proc. CoNLL}.

\bibitem[{Dosovitskiy et~al.(2020)Dosovitskiy, Beyer, Kolesnikov, Weissenborn,
  Zhai, Unterthiner, Dehghani, Minderer, Heigold, Gelly
  et~al.}]{dosovitskiy2020image}
Alexey Dosovitskiy, Lucas Beyer, Alexander Kolesnikov, Dirk Weissenborn,
  Xiaohua Zhai, Thomas Unterthiner, Mostafa Dehghani, Matthias Minderer, Georg
  Heigold, Sylvain Gelly, et~al. 2020.
\newblock An image is worth 16x16 words: Transformers for image recognition at
  scale.
\newblock In \emph{Proc. ICLR}.

\bibitem[{Fagarasan et~al.(2015)Fagarasan, Vecchi, and Clark}]{fagarasan2015}
Luana Fagarasan, Eva~Maria Vecchi, and Stephen Clark. 2015.
\newblock From distributional semantics to feature norms: Grounding semantic
  models in human perceptual data.
\newblock In \emph{Proc. {IWCS}}.

\bibitem[{Forbes et~al.(2019)Forbes, Holtzman, and Choi}]{forbes2019neural}
Maxwell Forbes, Ari Holtzman, and Yejin Choi. 2019.
\newblock Do neural language representations learn physical commonsense?
\newblock \emph{Proc. CogSci}.

\bibitem[{G{\"a}rdenfors(2000)}]{gardenfors2000}
Peter G{\"a}rdenfors. 2000.
\newblock \href {https://doi.org/10.7551/mitpress/2076.001.0001}
  {\emph{Conceptual Spaces: The Geometry of Thought}}.
\newblock The MIT Press.

\bibitem[{Hansen and Hebart(2022)}]{hansen2022}
Hannes Hansen and Martin~N. Hebart. 2022.
\newblock \href {https://arxiv.org/abs/2202.03753} {Semantic features of object
  concepts generated with {{GPT-3}}}.
\newblock In \emph{Proc. CogSci}.

\bibitem[{He et~al.(2022)He, Chen, Xie, Li, Doll{\'a}r, and
  Girshick}]{he2022masked}
Kaiming He, Xinlei Chen, Saining Xie, Yanghao Li, Piotr Doll{\'a}r, and Ross
  Girshick. 2022.
\newblock Masked autoencoders are scalable vision learners.
\newblock In \emph{Proc. CVPR}.

\bibitem[{Hebart et~al.(2023)Hebart, Contier, Teichmann, Rockter, Zheng,
  Kidder, Corriveau, {Vaziri-Pashkam}, and Baker}]{hebart2023}
Martin~N Hebart, Oliver Contier, Lina Teichmann, Adam~H Rockter, Charles~Y
  Zheng, Alexis Kidder, Anna Corriveau, Maryam {Vaziri-Pashkam}, and Chris~I
  Baker. 2023.
\newblock \href {https://doi.org/10.7554/eLife.82580} {{{THINGS-data}}, a
  multimodal collection of large-scale datasets for investigating object
  representations in human brain and behavior}.
\newblock \emph{eLife}, 12:e82580.

\bibitem[{Hebart et~al.(2019)Hebart, Dickter, Kidder, Kwok, Corriveau,
  Van~Wicklin, and Baker}]{hebart2019}
Martin~N. Hebart, Adam~H. Dickter, Alexis Kidder, Wan~Y. Kwok, Anna Corriveau,
  Caitlin Van~Wicklin, and Chris~I. Baker. 2019.
\newblock \href {https://doi.org/10.1371/journal.pone.0223792} {{{THINGS}}:
  {{A}} database of 1,854 object concepts and more than 26,000 naturalistic
  object images}.
\newblock \emph{PLOS ONE}, 14(10):e0223792.

\bibitem[{Hewitt and Liang(2019)}]{hewitt2019}
John Hewitt and Percy Liang. 2019.
\newblock \href {https://doi.org/10.18653/v1/D19-1275} {Designing and
  interpreting probes with control tasks}.
\newblock In \emph{Proc. EMNLP-IJCNLP}.

\bibitem[{Hill et~al.(2015)Hill, Reichart, and Korhonen}]{hill2015}
Felix Hill, Roi Reichart, and Anna Korhonen. 2015.
\newblock \href {https://doi.org/10.1162/COLI_a_00237} {Simlex-999: Evaluating
  semantic models with (genuine) similarity estimation}.
\newblock \emph{Computational Linguistics}, 41(4):665--695.

\bibitem[{Huh et~al.(2024)Huh, Cheung, Wang, and Isola}]{huh2024}
Minyoung Huh, Brian Cheung, Tongzhou Wang, and Phillip Isola. 2024.
\newblock The {P}latonic representation hypothesis.
\newblock In \emph{Proc. ICML}.

\bibitem[{Hupkes et~al.(2018)Hupkes, Veldhoen, and Zuidema}]{hupkes2018}
Dieuwke Hupkes, Sara Veldhoen, and Willem Zuidema. 2018.
\newblock \href {https://doi.org/10.1613/jair.1.11196} {Visualisation and
  'diagnostic classifiers' reveal how recurrent and recursive neural networks
  process hierarchical structure}.
\newblock \emph{Journal of Artificial Intelligence Research}, 61:907--926.

\bibitem[{Ivanova and Hofer(2020)}]{ivanova2020}
Anna~A. Ivanova and Matthias Hofer. 2020.
\newblock Linguistic overhypotheses in category learning: Explaining the label
  advantage effect.
\newblock In \emph{Proc. CogSci}.

\bibitem[{Li et~al.(2024)Li, Kementchedjhieva, Fierro, and
  S{\o}gaard}]{li2024c}
Jiaang Li, Yova Kementchedjhieva, Constanza Fierro, and Anders S{\o}gaard.
  2024.
\newblock \href {https://doi.org/10.1162/tacl_a_00698} {Do vision and language
  models share concepts? {A} vector space alignment study}.
\newblock \emph{TACL}, 12:1232--1249.

\bibitem[{Liu et~al.(2021)Liu, Bugliarello, Ponti, Reddy, Collier, and
  Elliott}]{liu2021visually}
Fangyu Liu, Emanuele Bugliarello, Edoardo~Maria Ponti, Siva Reddy, Nigel
  Collier, and Desmond Elliott. 2021.
\newblock Visually grounded reasoning across languages and cultures.
\newblock In \emph{Proc. EMNLP}.

\bibitem[{Liu et~al.(2024)Liu, Li, Li, and Lee}]{liu2024llava}
Haotian Liu, Chunyuan Li, Yuheng Li, and Yong~Jae Lee. 2024.
\newblock Improved baselines with visual instruction tuning.
\newblock In \emph{Proc. CVPR}.

\bibitem[{Liu et~al.(2022)Liu, Hu, Lin, Yao, Xie, Wei, Ning, Cao, Zhang, Dong
  et~al.}]{liu2022swin}
Ze~Liu, Han Hu, Yutong Lin, Zhuliang Yao, Zhenda Xie, Yixuan Wei, Jia Ning, Yue
  Cao, Zheng Zhang, Li~Dong, et~al. 2022.
\newblock Swin transformer v2: Scaling up capacity and resolution.
\newblock In \emph{Proc. CVPR}.

\bibitem[{Lucy and Gauthier(2017)}]{lucy2017}
Li~Lucy and Jon Gauthier. 2017.
\newblock \href {https://doi.org/10.18653/v1/W17-2810} {Are distributional
  representations ready for the real world? {E}valuating word vectors for
  grounded perceptual meaning}.
\newblock In \emph{Proceedings of the {{First Workshop}} on {{Language
  Grounding}} for {{Robotics}}}.

\bibitem[{Lupyan(2012)}]{lupyan2012}
Gary Lupyan. 2012.
\newblock \href {https://doi.org/10.3389/fpsyg.2012.00054} {Linguistically
  modulated perception and cognition: The label-feedback hypothesis}.
\newblock \emph{Frontiers in Psychology}, 3.

\bibitem[{Mahner et~al.(2024)Mahner, Muttenthaler, G{\"u}{\c{c}}l{\"u}, and
  Hebart}]{mahner2024}
Florian~P Mahner, Lukas Muttenthaler, Umut G{\"u}{\c{c}}l{\"u}, and Martin~N
  Hebart. 2024.
\newblock Dimensions underlying the representational alignment of deep neural
  networks with humans.
\newblock \emph{arXiv preprint arXiv:2406.19087}.

\bibitem[{McRae et~al.(2005)McRae, Cree, Seidenberg, and Mcnorgan}]{mcrae2005}
Ken McRae, George~S. Cree, Mark~S. Seidenberg, and Chris Mcnorgan. 2005.
\newblock \href {https://doi.org/10.3758/BF03192726} {Semantic feature
  production norms for a large set of living and nonliving things}.
\newblock \emph{Behavior Research Methods}, 37(4):547--559.

\bibitem[{Merullo et~al.(2023)Merullo, Castricato, Eickhoff, and
  Pavlick}]{merullo2023}
Jack Merullo, Louis Castricato, Carsten Eickhoff, and Ellie Pavlick. 2023.
\newblock \href {https://doi.org/10.48550/arXiv.2209.15162} {Linearly mapping
  from image to text space}.
\newblock In \emph{Proc. ICLR}.

\bibitem[{Mikolov et~al.(2018)Mikolov, Grave, Bojanowski, Puhrsch, and
  Joulin}]{mikolov2018advances}
Tomas Mikolov, Edouard Grave, Piotr Bojanowski, Christian Puhrsch, and Armand
  Joulin. 2018.
\newblock Advances in pre-training distributed word representations.
\newblock In \emph{Proc. LREC}.

\bibitem[{Misra et~al.(2023)Misra, Rayz, and Ettinger}]{misra2023}
Kanishka Misra, Julia Rayz, and Allyson Ettinger. 2023.
\newblock {COMPS}: Conceptual minimal pair sentences for testing robust
  property knowledge and its inheritance in pre-trained language models.
\newblock In \emph{Proc. EACL}.

\bibitem[{Misra et~al.(2022)Misra, Rayz, and Ettinger}]{misra2022}
Kanishka Misra, Julia~Taylor Rayz, and Allyson Ettinger. 2022.
\newblock A property induction framework for neural language models.
\newblock In \emph{Proc. CogSci}.

\bibitem[{Moutakanni et~al.(2024)Moutakanni, Oquab, Szafraniec, Vakalopoulou,
  and Bojanowski}]{moutakanni2024}
Th{\'e}o Moutakanni, Maxime Oquab, Marc Szafraniec, Maria Vakalopoulou, and
  Piotr Bojanowski. 2024.
\newblock You don't need domain-specific data augmentations when scaling
  self-supervised learning.
\newblock In \emph{Proc. NeurIPS}.

\bibitem[{Muttenthaler et~al.(2023)Muttenthaler, Dippel, Linhardt,
  Vandermeulen, and Kornblith}]{muttenthaler2023a}
Lukas Muttenthaler, Jonas Dippel, Lorenz Linhardt, Robert~A. Vandermeulen, and
  Simon Kornblith. 2023.
\newblock Human alignment of neural network representations.
\newblock In \emph{Proc. ICLR}.

\bibitem[{Nosofsky et~al.(2018)Nosofsky, Sanders, Meagher, and
  Douglas}]{nosofsky2018a}
Robert~M. Nosofsky, Craig~A. Sanders, Brian~J. Meagher, and Bruce~J. Douglas.
  2018.
\newblock \href {https://doi.org/10.3758/s13428-017-0884-8} {Toward the
  development of a feature-space representation for a complex natural category
  domain}.
\newblock \emph{Behavior Research Methods}, 50(2):530--556.

\bibitem[{Oquab et~al.(2024)Oquab, Darcet, Moutakanni, Vo, Szafraniec,
  Khalidov, Fernandez, Haziza, Massa, El-Nouby et~al.}]{oquabdinov2}
Maxime Oquab, Timoth{\'e}e Darcet, Th{\'e}o Moutakanni, Huy~V Vo, Marc
  Szafraniec, Vasil Khalidov, Pierre Fernandez, Daniel Haziza, Francisco Massa,
  Alaaeldin El-Nouby, et~al. 2024.
\newblock {DINO}v2: Learning robust visual features without supervision.
\newblock \emph{Transactions on Machine Learning Research}.

\bibitem[{Pennington et~al.(2014)Pennington, Socher, and
  Manning}]{pennington2014glove}
Jeffrey Pennington, Richard Socher, and Christopher~D Manning. 2014.
\newblock Glove: Global vectors for word representation.
\newblock In \emph{Proc. EMNLP}.

\bibitem[{Radford et~al.(2021)Radford, Kim, Hallacy, Ramesh, Goh, Agarwal,
  Sastry, Askell, Mishkin, Clark et~al.}]{radford2021learning}
Alec Radford, Jong~Wook Kim, Chris Hallacy, Aditya Ramesh, Gabriel Goh,
  Sandhini Agarwal, Girish Sastry, Amanda Askell, Pamela Mishkin, Jack Clark,
  et~al. 2021.
\newblock Learning transferable visual models from natural language
  supervision.
\newblock In \emph{Proc. ICML}.

\bibitem[{Rosch and Mervis(1975)}]{rosch1975}
Eleanor Rosch and Carolyn~B Mervis. 1975.
\newblock \href {https://doi.org/10.1016/0010-0285(75)90024-9} {Family
  resemblances: {{Studies}} in the internal structure of categories}.
\newblock \emph{Cognitive Psychology}, 7(4):573--605.

\bibitem[{Rubinstein et~al.(2015)Rubinstein, Levi, Schwartz, and
  Rappoport}]{rubinstein2015}
Dana Rubinstein, Effi Levi, Roy Schwartz, and Ari Rappoport. 2015.
\newblock \href {https://doi.org/10.3115/v1/P15-2119} {How well do
  distributional models capture different types of semantic knowledge?}
\newblock In \emph{Proc. ACL-IJCNLP}.

\bibitem[{Sharma et~al.(2018)Sharma, Ding, Goodman, and
  Soricut}]{sharma2018conceptual}
Piyush Sharma, Nan Ding, Sebastian Goodman, and Radu Soricut. 2018.
\newblock Conceptual {C}aptions: A cleaned, hypernymed, image alt-text dataset
  for automatic image captioning.
\newblock In \emph{Proc. ACL}.

\bibitem[{Silberer et~al.(2013)Silberer, Ferrari, and Lapata}]{silberer2013}
Carina Silberer, Vittorio Ferrari, and Mirella Lapata. 2013.
\newblock Models of semantic representation with visual attributes.
\newblock In \emph{Proc. ACL}.

\bibitem[{Sommerauer and Fokkens(2018)}]{sommerauer2018}
Pia Sommerauer and Antske Fokkens. 2018.
\newblock Firearms and tigers are dangerous, kitchen knives and zebras are not:
  Testing whether word embeddings can tell.
\newblock In \emph{Proc. {{EMNLP Workshop BlackboxNLP}}: {{Analyzing}} and
  {{Interpreting Neural Networks}} for {{NLP}}}.

\bibitem[{Speer et~al.(2017)Speer, Chin, and Havasi}]{speer2017conceptnet}
Robyn Speer, Joshua Chin, and Catherine Havasi. 2017.
\newblock {ConceptNet} 5.5: An open multilingual graph of general knowledge.
\newblock In \emph{Proc. AAAI}.

\bibitem[{Suresh et~al.(2024)Suresh, Huang, Mukherjee, and Rogers}]{suresh2024}
Siddharth Suresh, Wei-Chun Huang, Kushin Mukherjee, and Timothy~T Rogers. 2024.
\newblock Categories vs semantic features: What shapes the similarities people
  discern in photographs of objects?
\newblock In \emph{Proc. ICLR Workshop on Representational Alignment}.

\bibitem[{Team et~al.(2024)Team, Mesnard, Hardin, Dadashi, Bhupatiraju, Pathak,
  Sifre, Rivi{\`e}re, Kale, Love et~al.}]{team2024gemma}
Gemma Team, Thomas Mesnard, Cassidy Hardin, Robert Dadashi, Surya Bhupatiraju,
  Shreya Pathak, Laurent Sifre, Morgane Rivi{\`e}re, Mihir~Sanjay Kale,
  Juliette Love, et~al. 2024.
\newblock Gemma: Open models based on {G}emini research and technology.
\newblock \emph{arXiv preprint arXiv:2403.08295}.

\bibitem[{Tu et~al.(2022)Tu, Talebi, Zhang, Yang, Milanfar, Bovik, and
  Li}]{tu2022maxvit}
Zhengzhong Tu, Hossein Talebi, Han Zhang, Feng Yang, Peyman Milanfar, Alan
  Bovik, and Yinxiao Li. 2022.
\newblock Max{VIT}: Multi-axis vision transformer.
\newblock In \emph{Proc. ECCV}.

\bibitem[{Turton et~al.(2020)Turton, Vinson, and Smith}]{turton2020}
Jacob Turton, David Vinson, and Robert Smith. 2020.
\newblock Extrapolating binder style word embeddings to new words.
\newblock In \emph{Proc. Workshop on Linguistic and Neurocognitive Resources}.

\bibitem[{Utsumi(2020)}]{utsumi2018}
Akira Utsumi. 2020.
\newblock \href {https://doi.org/10.1111/cogs.12844} {Exploring what is encoded
  in distributional word vectors: A neurobiologically motivated analysis}.
\newblock \emph{Cognitive Science}, 44(6):e12844.

\bibitem[{Vuli{\'c} et~al.(2020)Vuli{\'c}, Ponti, Litschko, Glava{\v s}, and
  Korhonen}]{vulic2020}
Ivan Vuli{\'c}, Edoardo~Maria Ponti, Robert Litschko, Goran Glava{\v s}, and
  Anna Korhonen. 2020.
\newblock \href {https://doi.org/10.18653/v1/2020.emnlp-main.586} {Probing
  pretrained language models for lexical semantics}.
\newblock In \emph{Proc. EMNLP}.

\bibitem[{Waxman and Markow(1995)}]{waxman1995}
Sandra~R. Waxman and Dana~B. Markow. 1995.
\newblock \href {https://doi.org/10.1006/cogp.1995.1016} {Words as invitations
  to form categories: Evidence from 12- to 13-month-old infants}.
\newblock \emph{Cognitive Psychology}, 29(3):257--302.

\bibitem[{Xie et~al.(2022)Xie, Zhang, Cao, Lin, Bao, Yao, Dai, and
  Hu}]{xie2022simmim}
Zhenda Xie, Zheng Zhang, Yue Cao, Yutong Lin, Jianmin Bao, Zhuliang Yao,
  Qi~Dai, and Han Hu. 2022.
\newblock Sim{MIM}: A simple framework for masked image modeling.
\newblock In \emph{Proc. CVPR}.

\bibitem[{Zhai et~al.(2023)Zhai, Mustafa, Kolesnikov, and
  Beyer}]{zhai2023sigmoid}
Xiaohua Zhai, Basil Mustafa, Alexander Kolesnikov, and Lucas Beyer. 2023.
\newblock Sigmoid loss for language image pre-training.
\newblock In \emph{Proc. ICCV}.

\bibitem[{Zheng et~al.(2023)Zheng, Chiang, Sheng, Zhuang, Wu, Zhuang, Lin, Li,
  Li, Xing et~al.}]{zheng2023vicuna}
Lianmin Zheng, Wei-Lin Chiang, Ying Sheng, Siyuan Zhuang, Zhanghao Wu, Yonghao
  Zhuang, Zi~Lin, Zhuohan Li, Dacheng Li, Eric Xing, et~al. 2023.
\newblock Judging {LLM}-as-a-judge with {MT}-bench and {C}hatbot {A}rena.
\newblock In \emph{Proc. NeurIPS}.

\end{thebibliography}

\appendix

\section{Data Collection}
\label{sec:appendix-gpt4o-dataset}

\paragraph{Concept--attribute norm annotations.}
To obtain a complete representation of the \things concepts in terms of the (most frequent) attributes appearing in the McRae norms,
we asked GPT-4o (\texttt{gpt-4o-2024-08-06}) whether each norm is a valid trait of each concept;
Figure~\ref{fig:gpt4o-prompt-mcrae-x-things} shows the exact prompts.
Given 1854 concepts and 278 attributes, this yields over 515k queries.
We used the OpenAI Batch API for a the total cost of \$127.64.

\paragraph{Annotation validation.}
When extracting the annotations from the GPT-4o output, we observed that the format was not always consistent:
e.g.~the \texttt{valid} field was usually either \texttt{true} or \texttt{false},
but sometimes also \texttt{True}, \texttt{TRUE}, \texttt{yes}, \texttt{Yes}, \texttt{sometimes}, \texttt{False}, \texttt{no}, \texttt{No} (sometimes rendered as a string, sometimes as a literal);
sometimes the \texttt{valid} field also included explanations for the chosen answer or the concept definition;
sometimes the produced JSON used single quotes, sometimes double quotes.
In retrospect, many of these exceptions may have been prevented by a more precise prompting, but they were not apparent when testing at smaller scale.
To account for all these exceptions, we defined a custom parser that managed to extract a boolean value for each of the outputs.
The resulting data is available on the project's webpage.

\begin{figure*}
\begin{tcolorbox}
SYSTEM: ``You are asked to decide whether an attribute is a common trait of a concept (to follow). Please answer the request in JSON format with the following structure: \{`concept': CONCEPT, `attribute': ATTRIBUTE, `valid': ANSWER\}''

\vspace{1ex}

USER: "Is \{\texttt{attribute}\} a common trait of \{\texttt{concept}\}, in the sense of \{\texttt{concept\_definition}\}?"
\end{tcolorbox}
\caption{The prompt used to collect the \mxt dataset.}
\label{fig:gpt4o-prompt-mcrae-x-things}
\end{figure*}

\paragraph{Textual contexts.}
The best performance for contextualized language models depends on having a collection of sentences in which the concepts appear. 
In the absence of a large and naturally occurring dataset of such sentences, we prompted the GPT-4o API (\texttt{gpt4o-2024-08-06}) to collect the data. 
We also collected sentences with the addition constraint to avoid using any of the positively-labelled semantic norms for a given concept. 
(This was in order to reduce the chance that the resulting embedding literally included features about the expected norm.) 
Figure~\ref{fig:gpt4o-sentences-prompt} shows the prompts used.
The total cost of collecting the sentences was \$26.24.

\begin{figure*}
\begin{tcolorbox}
SYSTEM: ``You are asked to write \{\texttt{num}\} short sentences about a word (to follow). Answer the request by returning a list of numbered sentences, 1--\{\texttt{num}\}.''

\vspace{1ex}

USER: ``Write \{\texttt{num}\} short sentences about \{\texttt{concept}\}. You must use \{\texttt{concept}\} as a noun in each sentence.''
\end{tcolorbox}

\begin{tcolorbox}
SYSTEM: ``You are asked to write \{\texttt{num}\} short sentences about a word (to follow). Answer the request by returning a list of numbered sentences, 1--\{\texttt{num}\}.''

\vspace{1ex}

USER: ``Write \{\texttt{num}\} short sentences about \{\texttt{concept}\}. You must use \{\texttt{concept}\} as a noun in each sentence. Try to avoid using the following phrases in any of the sentences: \{\texttt{positive}\_\texttt{attributes}\}''
\end{tcolorbox}

\caption{The prompts used to collect sentence contexts for each concept in the \things dataset. Top: Unconstrained prompt; Bottom: Constrained prompt.
The constraint tries to prevent GPT4o from mentioning the attributes already associated with a concept.}
\label{fig:gpt4o-sentences-prompt}
\end{figure*}

\section{Model Details}
\label{app:model-details}

\begin{table}
    \centering
    \small
    \setlength{\tabcolsep}{4pt}
    \begin{tabular}{ll}
    \toprule
    ViT-MAE & \texttt{facebook/vit-mae-large} \\
    Max ViT 1K & \texttt{maxvit\_large\_tf\_384.in1k} \\
    Max ViT 21K & \texttt{maxvit\_large\_tf\_224.in21k} \\
    DINOv2 & \texttt{facebook/dinov2-large} \\
    Swin-V2 & \texttt{swinv2\_large\_window12\_192.ms\_in22k} \\
    \midrule
    LLaVA-1.5 & \texttt{llava-hf/llava-1.5-7b-hf} \\
    Qwen2.5-VL & \texttt{Qwen/Qwen2.5-VL-3B-Instruct} \\
    CLIP & \texttt{openai/clip-vit-large-patch14} \\
    PaliGemma & \texttt{google/paligemma-3b-mix-224} \\
    SigLIP & \texttt{google/siglip-so400m-patch14-224} \\
    \midrule
    GLoVe & \texttt{glove-840b-300d} \\
    DeBERTa~v3 & \texttt{deberta-v3} \\
    Gemma & \texttt{google/gemma-2b} \\
    \bottomrule
    \end{tabular}
    \caption{Precise names of the models used in this paper.}
    \label{tab:model_names}
\end{table}

For reproducibility, Table~\ref{tab:model_names} shows the exact model versions used in the experiments.

\section{Further Results}
\label{app:further-results}

\paragraph{Detailed results.}
Table~\ref{tab:main-results-full} presents the results in terms of precision, recall, raw F$_1$, and F$_1$ selectivity scores
for the \mxt dataset and median-binarised Binder dataset.
On the original Binder dataset, we report root mean squared and mean absolute errors.

\begin{table*}
    \centering
    \setlength{\tabcolsep}{5pt}
\begin{tabularx}{\linewidth}{X rrrr rrrr rr}
\toprule
& \multicolumn{4}{c}{\mxt}
& \multicolumn{4}{c}{Binder (binarised)} 
& \multicolumn{2}{c}{Binder} 
\\
\cmidrule(lr){2-5}
\cmidrule(lr){6-9} 
\cmidrule(lr){10-11} 
Model
& P $\uparrow$ & R $\uparrow$ & F$_1$ $\uparrow$ & F$_1$ sel $\uparrow$
& P $\uparrow$ & R $\uparrow$ & F$_1$ $\uparrow$ & F$_1$ sel $\uparrow$
& RMSE $\downarrow$ & MAE $\downarrow$ \\
\midrule
\multicolumn{11}{l}{\textit{Vision models}} \\
Random SigLIP    & 26.2     & 28.0     & 26.8     & 15.4     & 60.6     & 60.3     & 59.8     & 9.3      & 1.43     & 1.12 \\
ViT-MAE          & 49.6     & 46.1     & 47.0     & 35.6     & 70.0     & 70.0     & 69.4     & 18.8     & 0.94     & 0.73 \\
Max ViT (IN-1K)  & 38.7     & 44.1     & 40.4     & 29.0     & 62.2     & 61.0     & 61.0     & 10.4     & 1.37     & 1.07 \\
Max ViT (IN-21K) & 63.5     & 50.3     & 54.7     & 43.3     & 71.6     & 73.6     & 72.0     & 21.5     & 0.84     & 0.65 \\
DINOv2           & 59.8     & \bf 54.0 & 55.9     & 44.5     & 73.8     & 73.7     & 73.2     & 22.7     & 0.80     & 0.61 \\
Swin-V2          & \bf 67.3 & 53.8     & \bf 58.4 & \bf 47.0 & \bf 74.8 & \bf 75.2 & \bf 74.5 & \bf 23.9 & \bf 0.74 & \bf 0.55 \\
\midrule
\multicolumn{11}{l}{\textit{Multimodal vision models}} \\
LLaVA-1.5        & 59.1     & 55.5     & 56.4     & 45.0     & 74.6     & 74.0     & 73.8     & 23.2     & 0.83     & 0.64 \\
Qwen2.5-VL       & 62.0     & 56.6     & 58.2     & 46.8     & 75.4     & 75.0     & 74.7     & 24.1     & 0.79     & 0.61 \\
CLIP (image)     & 63.5     & 58.1     & 59.8     & 48.4     & \bf 77.0 & \bf 76.2 & \bf 76.1 & \bf 25.5 & 0.74     & 0.56 \\
PaliGemma        & 67.3     & 58.2     & 61.3     & 49.9     & 76.0     & 76.1     & 75.5     & 25.0     & 0.73     & 0.55 \\
SigLIP           & \bf 67.5 & \bf 58.4 & \bf 61.5 & \bf 50.1 & 76.8     & 76.0     & 75.8     & 25.2     & \bf 0.71 & \bf 0.53 \\
\midrule
\multicolumn{11}{l}{\textit{Language models}} \\
GloVe 840B       & 51.9     & 51.1     & 50.5     & 39.1     & 74.6     & 74.1     & 73.9     & 23.3     & 0.89     & 0.69 \\
FastText         & 55.1     & 50.7     & 51.6     & 40.2     & 74.0     & 74.1     & 73.5     & 22.9     & 0.91     & 0.71 \\
Numberbatch      & 59.6     & 54.0     & 55.5     & 44.1     & 75.0     & 75.0     & 74.5     & 23.9     & 0.83     & 0.65 \\
CLIP (text)      & 60.2     & 51.7     & 54.4     & 43.0     & 73.2     & 72.7     & 72.5     & 21.9     & 0.81     & 0.63 \\
DeBERTa v3       & 64.2     & 53.2     & 56.9     & 45.5     & 76.9     & 76.1     & 75.9     & 25.3     & 0.68     & 0.52 \\
Gemma            & \bf 68.7 & \bf 57.2 & \bf 61.2 & \bf 49.8 & \bf 77.1 & \bf 76.5 & \bf 76.3 & \bf 25.7 & \bf 0.67 & \bf 0.51 \\
\bottomrule
\end{tabularx}
    \caption{%
        Detailed results, in terms of precision (P), recall (R), F$_1$ score (F$_1$) and F$_1$ selectivity score (F$_1$ sel),
        of concept norm linear probes on the \mxt and binarised Binder datasets.
        On the original Binder dataset we report root mean squared error (RMSE) and mean absolute error (MAE).
    }
    \label{tab:main-results-full}
\end{table*}

\begin{figure*}
    \includegraphics[width=\linewidth]{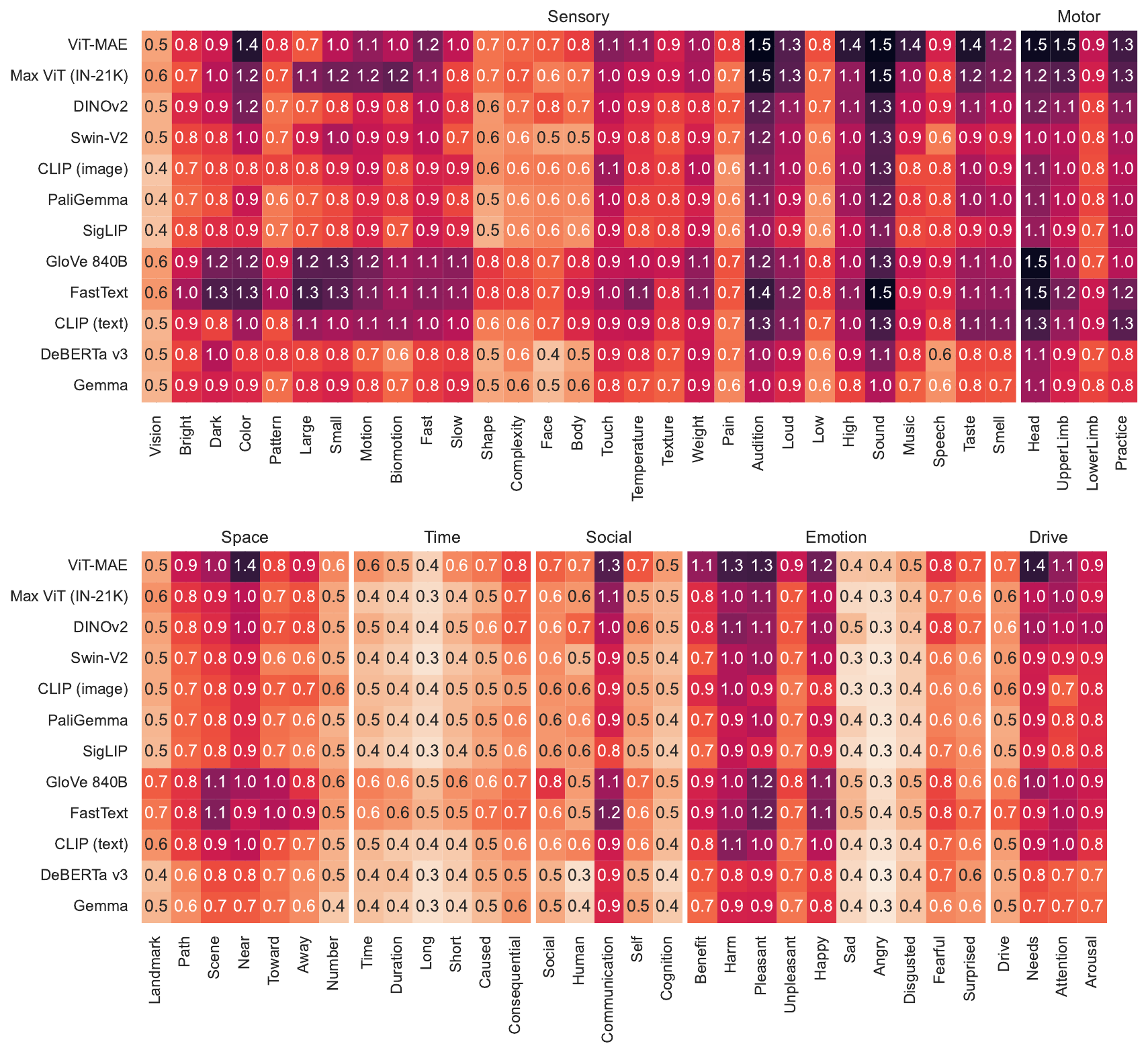}
    \caption{%
    Per-attribute RMSE on Binder attribute ratings, across models. Lower is better.
    }
    \label{fig:binder-norms-results}
\end{figure*}

\begin{figure*}
    \includegraphics[width=\linewidth]{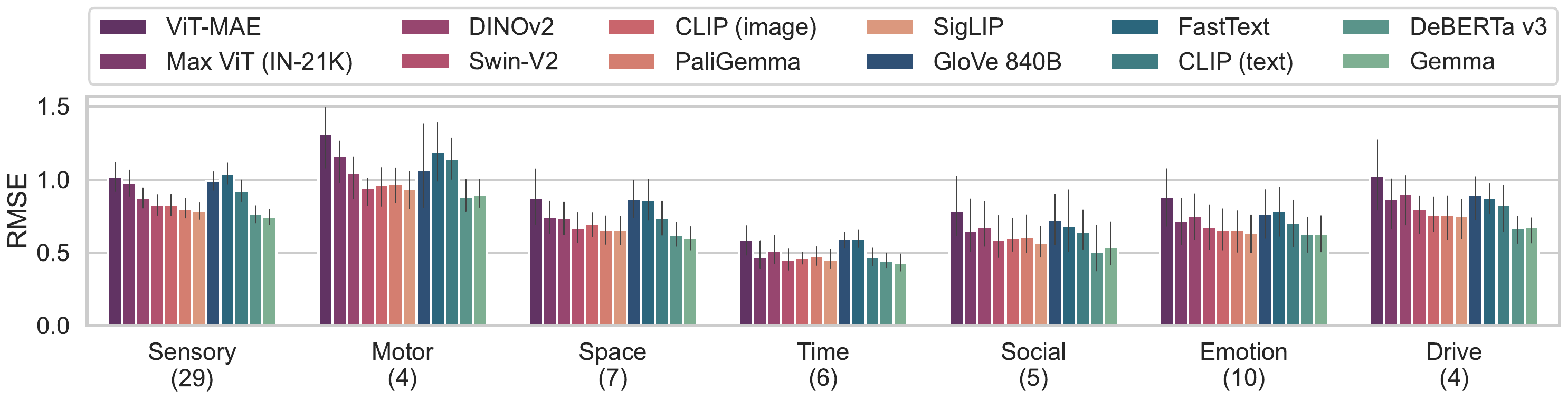}
    \caption{%
    Results (RMSE) aggregated over attribute domain on the Binder data (note: lower is better).
    The number below each domain indicates the number of attributes belonging to that domain.
    The error bars denote 95\% confidence intervals using bootstrapping.
    Vision models are in reddish colours, while language models are in greenish colours.}
    \label{fig:binder-norms-results-by-type}
\end{figure*}

\paragraph{Per-attribute results on Binder.}
Figure~\ref{fig:binder-norms-results} presents the detailed results on each of the 67 attributes from the Binder dataset.
Figure~\ref{fig:binder-norms-results-by-type} shows the results aggregated per attribute type (7 types).
We see that the auditory attributes (\attribute{audition, loud, sound}) are the most difficult.
Distinguishing between positively and negatively associated concepts (\attribute{Benefit, Harm, Pleasant, Unpleasant, Happy}) is also surprisingly difficult.
Interestingly, attributes to do with Time and negative Emotions (\attribute{sad, angry, disgusted}) are relatively easy for most models.
Attributes that have directly to do with the human body (\attribute{Face, Body, Self, Human}) are also fairly easy.

\section{Qualitative Results}
\label{sec:qualitiative results}

\begin{figure*} 
    \centering
    \newcommand{\correct}[1]{{\color{green!40!black} #1}}
\newcommand{\wrong}[1]{{\color{red!80!black} #1}}
\newcommand{\cyes}{\textbf{+}}
\newcommand{\cno}{\textbf{--}}
\newcommand{\yes}{{\scriptsize\checkmark}}
\newcommand{\no}{$\bullet$}
\newcommand{\TP}{{\color{green!40!black} TP}}
\newcommand{\TN}{{\color{gray} TN}}
\newcommand{\FP}{{\color{orange!80!black} FP}}
\newcommand{\FN}{{\color{red!80!black} FN}}
\newcommand{\ww}{0.08\textwidth}
\footnotesize
\setlength{\tabcolsep}{4pt}
\begin{tabular}{lr cc cc cc cc cc}
Model & F1 sel. & \multicolumn{10}{c}{Five random samples per attribute and their predictions} \\
\midrule
& & \multicolumn{10}{c}{\attribute{has 4 legs} (visual: form \& surface)} \\
 &  & \concept{dog} & \cyes & \concept{stool} & \cyes & \concept{tablecloth} & \cno & \concept{altar} & \cno & \concept{kangaroo} & \cno \\
Swin-V2 & 78.5 & \multirow{4}{*}{\includegraphics[width=\ww]{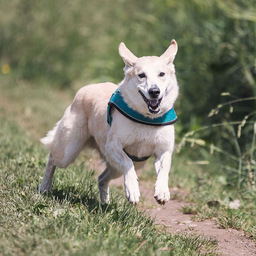}} & \correct{\yes} & \multirow{4}{*}{\includegraphics[width=\ww]{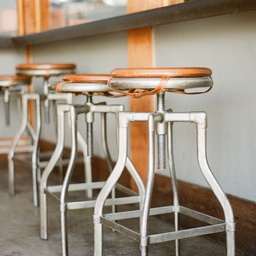}} & \correct{\yes} & \multirow{4}{*}{\includegraphics[width=\ww]{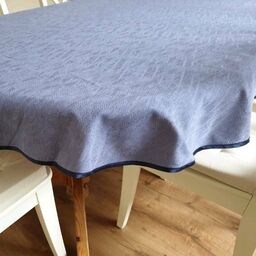}} & \wrong{\yes} & \multirow{4}{*}{\includegraphics[width=\ww]{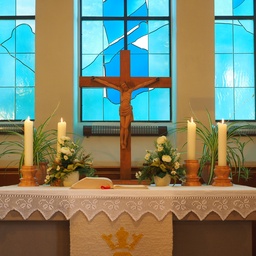}} & \correct{\no} & \multirow{4}{*}{\includegraphics[width=\ww]{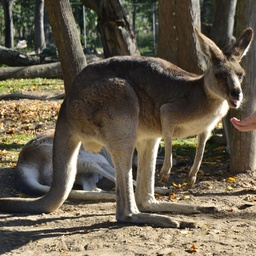}} & \wrong{\yes} \\
Gemma & 75.7 &  & \correct{\yes} &  & \wrong{\no} &  & \correct{\no} &  & \wrong{\yes} &  & \wrong{\yes} \\
CLIP (image) & 76.6 &  & \correct{\yes} &  & \correct{\yes} &  & \wrong{\yes} &  & \wrong{\yes} &  & \wrong{\yes} \\
CLIP (text) & 71.6 &  & \correct{\yes} &  & \correct{\yes} &  & \wrong{\yes} &  & \wrong{\yes} &  & \wrong{\yes} \\
[0.1cm]
\midrule
& & \multicolumn{10}{c}{\attribute{made of wood} (visual: form \& surface)} \\
 &  & \concept{axe} & \cyes & \concept{ski} & \cyes & \concept{bow3} & \cno & \concept{puppet} & \cno & \concept{cardboard} & \cno \\
Swin-V2 & 46.1 & \multirow{4}{*}{\includegraphics[width=\ww]{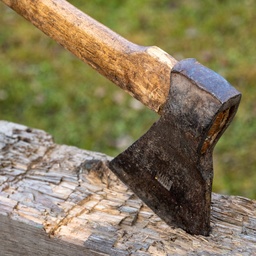}} & \correct{\yes} & \multirow{4}{*}{\includegraphics[width=\ww]{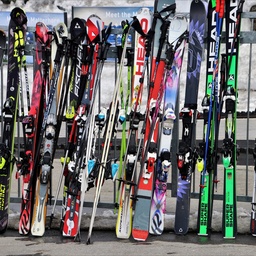}} & \wrong{\no} & \multirow{4}{*}{\includegraphics[width=\ww]{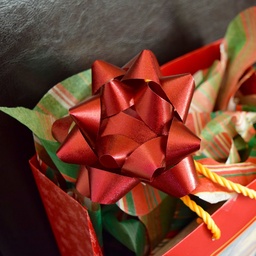}} & \correct{\no} & \multirow{4}{*}{\includegraphics[width=\ww]{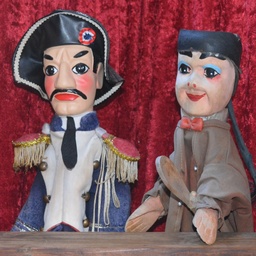}} & \correct{\no} & \multirow{4}{*}{\includegraphics[width=\ww]{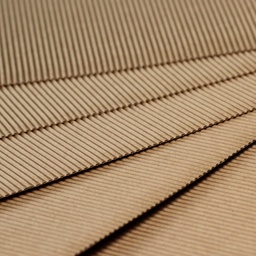}} & \wrong{\yes} \\
Gemma & 49.7 &  & \correct{\yes} &  & \wrong{\no} &  & \correct{\no} &  & \wrong{\yes} &  & \wrong{\yes} \\
CLIP (image) & 47.8 &  & \correct{\yes} &  & \wrong{\no} &  & \correct{\no} &  & \correct{\no} &  & \correct{\no} \\
CLIP (text) & 43.8 &  & \correct{\yes} &  & \wrong{\no} &  & \wrong{\yes} &  & \wrong{\yes} &  & \wrong{\yes} \\
[0.1cm]
\midrule
& & \multicolumn{10}{c}{\attribute{is dangerous} (encyclopaedic)} \\
 &  & \concept{dynamite} & \cyes & \concept{bison} & \cyes & \concept{razor} & \cyes & \concept{corkscrew} & \cno & \concept{tattoo} & \cno \\
Swin-V2 & 38.7 & \multirow{4}{*}{\includegraphics[width=\ww]{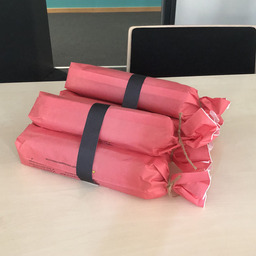}} & \correct{\yes} & \multirow{4}{*}{\includegraphics[width=\ww]{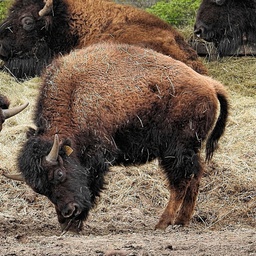}} & \correct{\yes} & \multirow{4}{*}{\includegraphics[width=\ww]{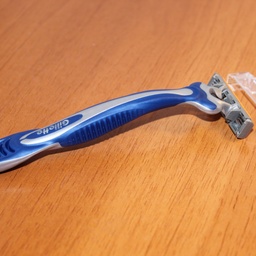}} & \wrong{\no} & \multirow{4}{*}{\includegraphics[width=\ww]{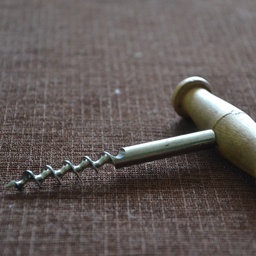}} & \wrong{\yes} & \multirow{4}{*}{\includegraphics[width=\ww]{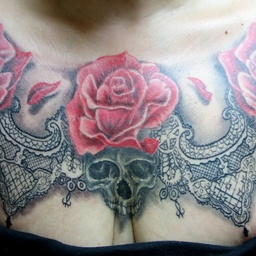}} & \correct{\no} \\
Gemma & 51.0 &  & \correct{\yes} &  & \wrong{\no} &  & \correct{\yes} &  & \correct{\no} &  & \wrong{\yes} \\
CLIP (image) & 44.8 &  & \correct{\yes} &  & \correct{\yes} &  & \wrong{\no} &  & \correct{\no} &  & \correct{\no} \\
CLIP (text) & 38.9 &  & \correct{\yes} &  & \correct{\yes} &  & \correct{\yes} &  & \correct{\no} &  & \correct{\no} \\
[0.1cm]
\midrule
& & \multicolumn{10}{c}{\attribute{tastes sweet} (taste)} \\
 &  & \concept{plum} & \cyes & \concept{raisin} & \cyes & \concept{cake mix} & \cyes & \concept{tomato sauce} & \cno & \concept{crystal1} & \cno \\
Swin-V2 & 72.9 & \multirow{4}{*}{\includegraphics[width=\ww]{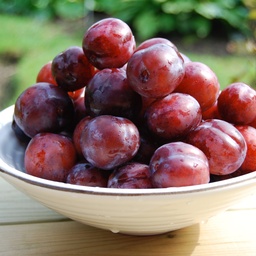}} & \correct{\yes} & \multirow{4}{*}{\includegraphics[width=\ww]{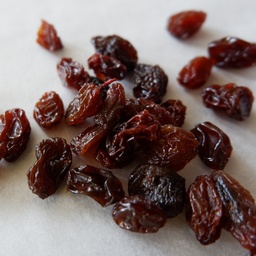}} & \correct{\yes} & \multirow{4}{*}{\includegraphics[width=\ww]{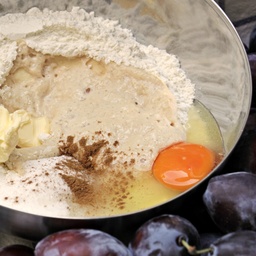}} & \wrong{\no} & \multirow{4}{*}{\includegraphics[width=\ww]{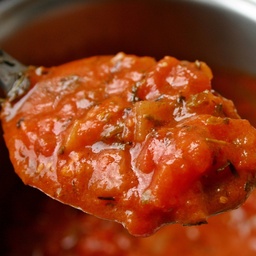}} & \correct{\no} & \multirow{4}{*}{\includegraphics[width=\ww]{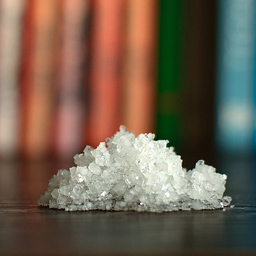}} & \wrong{\yes} \\
Gemma & 71.8 &  & \correct{\yes} &  & \wrong{\no} &  & \correct{\yes} &  & \wrong{\yes} &  & \correct{\no} \\
CLIP (image) & 72.9 &  & \correct{\yes} &  & \correct{\yes} &  & \wrong{\no} &  & \correct{\no} &  & \correct{\no} \\
CLIP (text) & 59.8 &  & \correct{\yes} &  & \correct{\yes} &  & \correct{\yes} &  & \correct{\no} &  & \correct{\no} \\
[0.1cm]
\bottomrule
\end{tabular}
    \caption{%
        Five random predictions of linear probes trained on four attributes.
        Positive concepts are indicated by \cyes, negative concepts by \cno.
        The linear probes are trained on embeddings from one of the four models: Swin-V2, Gemma, CLIP image and text encoders.
        If a model predicts a concept as having the attribute,
        we indicate this by \yes; otherwise we use \no.
        The correctness of the prediction is colour-coded:
        green for a correct prediction, red for an incorrect one.
        In the second column, we show the F1 selectivity (\%) for the each of the models and attributes.
    }
    \label{fig:qualitative-results}
\end{figure*}

In Figure~\ref{fig:qualitative-results} we show results at the level of attributes and concepts.
The results are four attributes (\attribute{has 4 legs}, \attribute{made of wood}, \attribute{is dangerous}, \attribute{tastes sweet}),
and for each we show five random samples (concepts).
For each sample we provide, the prediction using the same model selection as at the end of Section~\ref{subsec:main-results}: that is,
the best vision-only model (Swin-V2),
the best language-only model (Gemma),
and the language-and-vision models (CLIP image and CLIP text).
Note that the models ingest the concept samples differently:
the vision models average embeddings over multiple images, Gemma uses contextual sentences;
so the images and concept word in Figure~\ref{fig:qualitative-results} are shown for illustrative purposes.

For the attribute \attribute{has 4 legs} we see that the vision-based models (Swin-V2 and CLIP-image) label \concept{tablecloth} as positive, likely due to visual co-occurrence with \concept{table}.
All models struggle with the difficult cases of \concept{kangaroo}, predicted as \attribute{having 4 legs}, and \concept{ski}, predicted as not \attribute{made of wood}.
Some concept--attribute pairs are arguably ambiguous---is a \concept{corkscrew} \attribute{dangerous}? is a \concept{tomato sauce} \attribute{sweet}?---resulting in disagreements between models.

\section{Failures in Extracting Contextualized Textual Representations}\label{sec:appendix-textual_contexts}

\begin{table*}
    
\setlength{\tabcolsep}{5pt}
\begin{tabularx}{\linewidth}{lXl c r@{\hspace{0.1cm}}l rrrr}
    \toprule
    & & & & & & \multicolumn{4}{c}{\mxt}\\
    \cmidrule(lr){7-10} 
    & Model & Input & Seq. & \multicolumn{2}{c}{Layer} & P & R & F$_1$ & F$_1$ sel \\
    \midrule
    A & Gemma   & word                    & mean & 0     & (emb)     & 43.2 & 25.3 & 30.3 & 18.8 \\
    B & Gemma   & word (space)            & mean & 0     & (emb)     & 58.3 & 37.9 & 44.2 & 32.8 \\
    C & Gemma   & sentences (10)          & mean & 1     &           & 61.2 & 41.8 & 47.9 & 36.5 \\
    D & Gemma   & sentences (10)          & mean & 18    & (last)    & 63.8 & 52.4 & 56.3 & 44.9 \\
    E & Gemma   & sentences (10)          & last & 18    & (last)    & 66.5 & 56.8 & 60.2 & 48.8 \\
    F & Gemma   & sentences (10)          & mean & 0--6  &           & 62.2 & 46.3 & 51.5 & 40.1 \\
    G & Gemma   & sentences (10)          & mean & 0--9  &           & 62.3 & 48.7 & 53.2 & 41.8 \\
    H & Gemma   & sentences (10)          & mean & 9--18 &           & 65.9 & 53.9 & 58.0 & 46.6 \\
    I & Gemma   & sentences (10)          & last & 9--18 &           & 68.7 & 57.2 & 61.2 & 49.8 \\
    J & Gemma   & sentences (50)          & mean & 18    & (last)    & 62.7 & 52.1 & 55.8 & 44.4 \\
    K & Gemma   & sentences (50, constr.) & mean & 18    & (last)    & 62.1 & 51.6 & 55.2 & 43.8 \\
    \midrule                                                          
    L & DeBERTa v3 & sentences (10)          & mean & 12    & (last) & 43.9 & 42.9 & 42.8 & 31.4 \\
    M & DeBERTa v3 & sentences (10)          & mean & 0--4  &        & 62.9 & 51.6 & 55.3 & 43.9 \\
    N & DeBERTa v3 & sentences (10)          & mean & 0--6  &        & 64.2 & 53.2 & 56.9 & 45.5 \\
    \midrule                                                          
    O & GPT2    & sentences (10)          & mean & 12    & (last)    & 45.4 & 41.1 & 42.4 & 31.0 \\
    \midrule                                                          
    P & BERT base uncased & sentences (10)          & mean & 0--4  & & 48.9 & 41.1 & 43.5 & 32.0 \\
    Q & BERT base uncased & sentences (10)          & mean & 0--6  & & 50.9 & 42.7 & 45.2 & 33.8 \\
    \bottomrule
\end{tabularx}
    \caption{The effects of
    input (isolated concept word or contextual sentences),
    sequence pooling (mean or last token), and
    layer (individual layer or averaged over a range of layers)
    for the contextualised language models.}
    \label{tab:language-models-results}
\end{table*}

Concept representations can, in principle, be extracted from any language model using just the surface-form of the concept label token(s). 
Here, we report a collection of negative results for this seemingly simple task using contextual language models. 
Table~\ref{tab:language-models-results} presents the complete results of our endeavours.
Initial experiments with the Gemma-2B language model focused on using only the static embedding layer, which resulted in complete failure to train meaningful probes (\textbf{A}).
Closer inspection revealed that the Gemma-2B tokenizer tokenizes single word inputs differently from words appearing in a sentence (i.e.,~words preceded by a space): \texttt{<bos>aardvark}$\rightarrow$\{\texttt{aard}, \texttt{vark}\} instead of \{\texttt{\_aard}, \texttt{vark}\}.
Using the within-sentence (space-prepended) tokenization, performance improved but was still lower than expected (\textbf{B}).
Nevertheless, this approach was still substantially below the performance that we expected.
Following \citet{bommasani2020}, we decided to collect contextualized sentence representations over a set of textual contexts for each concept.
We collected 50 sentences from the GPT-4o API for each context (see Appendix~\ref{sec:appendix-gpt4o-dataset} for details).
These per sentence embeddings are averaged over multiple sentences, analogous to averaging the embeddings over multiple image instances.
This greatly improved performance compared to using the embedding layer (\textbf{C}), and extracting the representation from the last later further improved performance (\textbf{D}).
Another improvement was obtained by extracting the representation from the final subword token of a concept, i.e. \texttt{vark} in the tokenization of \texttt{aardvark} (\textbf{E}), and the final improvement involved extracting the representation as an average over multiple Transformer layers (\textbf{I}).
The representations obtained from 50 sentences did not improve performance (\textbf{J}).
Performance was slightly reduced using the contexts generated with the semantic norm constraints (\textbf{K}), indicating the model could use information from context sentences for this task.
With this methodology fixed, we quickly found better representations for the DeBERTa~v3 language encoder (\textbf{N}), and confirmed that this would also result in marginal improvements for BERT (\textbf{Q}).
We also report results for BERT base (uncased) and GPT-2 for completeness.
We find that BERT base (uncased) performs much worse than DeBERTa~v3 in similar conditions (\textbf{N} vs \textbf{Q}), and that GPT-2 also performs much worse than Gemma (\textbf{O} vs \textbf{D}).
Given these findings, we do not include BERT or GPT-2 in our main results.

\end{document}